# The 3rd International Planning Competition: Results and Analysis

**Derek Long**                                              DEREK.LONG@CIS.STRATH.AC.UK
**Maria Fox**                                               MARIA.FOX@CIS.STRATH.AC.UK
*Department of Computer and Information Sciences*
*University of Strathclyde, Glasgow, UK*

## Abstract

This paper reports the outcome of the third in the series of biennial international planning competitions, held in association with the International Conference on AI Planning and Scheduling (AIPS) in 2002. In addition to describing the domains, the planners and the objectives of the competition, the paper includes analysis of the results. The results are analysed from several perspectives, in order to address the questions of comparative performance between planners, comparative difficulty of domains, the degree of agreement between planners about the relative difficulty of individual problem instances and the question of how well planners scale relative to one another over increasingly difficult problems. The paper addresses these questions through statistical analysis of the raw results of the competition, in order to determine which results can be considered to be adequately supported by the data. The paper concludes with a discussion of some challenges for the future of the competition series.

## 1. Introduction

Beginning in 1998 the international planning community has held a biennial event to support the direct comparison of planning systems on a changing collection of benchmark planning problems. The benefits of this series of events have been significant: over five years, planning systems have been developed that are capable of solving large and complex problems, using richly expressive domain models and meeting advanced demands on the structure and quality of solutions. The competition series has inspired many advances in the planning research community as well as an increasingly empirical methodology and a growing interest in the application of planners to real problems.

In this paper we describe the structure, objectives and outcomes of the third competition, which took place in Toulouse in 2002. The competition was colocated with the AI Planning and Scheduling (AIPS) conference. At that conference a brief report was presented of some of the results achieved by the participating planners. We begin by presenting an overview of the main results as presented at the conference, showing the number of problems attempted and solved by each planner and identifying the competition prize-winners. As in previous years the competition resulted in the collection of a large data set comprising data points for several different domains. A certain comparative understanding can be obtained by examining the data for the individual domains, but conclusions drawn on this basis cannot be generalised across domains. One of the goals of this paper is to try to reveal some insights that cross the boundaries of the domains and allow some general questions to be answered. These include: which planners reveal the most consistent, stable performance across domains? What benefit is obtained by exploiting hand-coded control knowledge? Is there any general agreement over what makes a planning problem hard? Are particular planning approaches best suited to particular kinds of problem domains?

The accepted scientific methodology for addressing such questions is to frame precise hypotheses *prior* to the collection of the data sets in order to control for any extraneous variables that might distort the reality with respect to these questions. To date this has not been the practice of the planning community with respect to the competitions. In the third competition we proceeded, as in previous years, by collecting data prior to detailed consideration of the specific questions we wished to answer. The community has not yet agreed that the primary role of the competition is to provide a carefully crafted platform for the scientific investigation of planners: indeed, its main roles so far have been to motivate researchers in the field, to identify new research goals and thereby push forward the research horizons, and to publicise progress to the wider community. However, because competitions have winners there is a natural tendency to draw conclusions from the competition data sets about the state of the art. If these conclusions are not scientifically supported they can be misleading and even erroneous. Therefore there is an argument for trying to combine the two objectives, although admittedly there is a tension between them that might make it difficult to combine them successfully.

Because of the way in which the planning competitions are currently conducted, the analyses we describe in this paper are post hoc. We conducted all of our analyses on the data collected during the competition period: we did not run any further experiments after the competition because the participants felt that it was important that the data they submitted during the competition should comprise the evidence on which they were judged. We have identified a number of analyses which we think provide interesting information to the planning community and, in the following sections, we explore each theme in as rigorous a way as possible within the constraints of the data we have at our disposal. It has been difficult to work with a fixed data set that was collected without precise experimental questions in mind, and we were unable to test many of the hypotheses that occurred to us during our analyses because the data was inappropriate or incomplete. However, despite the limitations of the data set we believe we have been able to pose and answer some important questions about the comparative performances revealed in the competition. We phrase the objectives of our analyses in terms of null and alternative hyptheses because this is the standard approach when applying statistical tests. Our approach was partly inspired by the earlier work of Howe and Dahlman (2002). Their work raised the standard to which evaluation of planners should be done and compels the planning community to decide whether future planning competitions should be conducted in a way that supports the goals of scientific investigation of progress in the field.

The rest of this paper is organised as follows. We begin by discussing the context — the competition series itself and the general form of the third competition, including the domains, the competitors and the specific challenges raised. We then briefly summarise the results of the competition before embarking on a detailed post hoc analysis of the competition results. This analysis investigates the relative performances of the planners, the relative difficulties of the problem sets used and the relative scaling behaviours of the competitors across domains and levels. We provide an appendix in which we present summaries of the competing planners and details of the domains used.

## 2. The International Planning Competition Series

The first event was held in conjunction with the fourth international Artificial Intelligence Planning and Scheduling conference (AIPS'98). It was organised by Drew McDermott, together with a committee of senior members of the research community (McDermott, 2000). The event took some time to organise, with evolving agreement on the form of the event, the kinds of planning systems that should be compared, the basis of comparison and so on. The final event became a direct comparison of 5 STRIPS-based planners, with two of the planners also attempting an extended ADL-based language (McDermott, 2000; Long, 2000). The systems included three Graphplan-based systems, one forward heuristic search system and one planning-as-satisfiability SATsolver planner. A very important outcome of the first competition was the adoption of PDDL (McDermott & the AIPS'98 Planning Competition Committee, 1998) as a common representation language for planning.

Although the opportunity was offered for competitors to hand-code control knowledge for their planners, in fact all of the planners were fully-automated and ran on the problem instances without any priming. The entire event was staged at the conference over a period of some four days, involving intensive sessions of generating and checking solutions and attempting to evaluate the results. One idea that was tried, but that turned out to be problematic in practice, was to score the planners' performances using a function that attempted to take into account the time taken to generate a plan, the length of the plan and the relative performance of all of the competitors on the problems. For example, a planner that produced a plan faster than all of its competitors would be rewarded based on how much faster it was than the average for the problem. This attempt to score planners using a one-dimensional measure proved difficult, with counter-intuitive results in certain cases. In the end it was abandoned in favour of two dimensions: length of plan and time taken to produce it. This decision indicates that, even for only five systems and a relatively small set of problems, it is impossible to make unequivocal decisions about which system is best. Nevertheless, the community can (and did) learn much from the data that is gathered, offering a variety of interpretations of the data, but ultimately being inspired to improve on it in every way possible.

In the second competition, chaired by Fahiem Bacchus, 17 planners competed. The increase in participation and the ambitions for larger scale testing required that the event be spread over a much longer period. In fact, testing was spread over a couple of months, with only one final test being carried out at the conference site (AIPS'00 in Breckenridge). In the second competition there was a more formal split between systems, with a small number using hand-coded control knowledge and others being fully-automated. There was also a split between STRIPS and ADL capable systems. The larger number of competitors included a wider range of approaches: as well as Graphplan-based systems, forward heuristic search and a SATsolver, there were several planners based on model-checking approaches using BDDs, and one using planning-by-rewriting. Again, it proved difficult to compare planners unequivocally, but several important observations could be made: the advantages of hand-coded control rules in most domains could be seen clearly (as would be expected), although there remained an important question about the difficulty of generating and writing the rules. Of the fully-automated planners, the forward heuristic search approach proved to be particularly successful, dominating performance in most domains. Pure Graphplan-based planning seemed to have reached its zenith between the first two competitions and no longer appeared competitive.

The third competition (and most recent at the time of writing) was held in association with AIPS'02 at Toulouse. Fourteen planners participated. The primary objective of the competition was to help to push forward research into temporal and resource-intensive planning. Extensions were made to PDDL to support the modelling of temporal and numeric domain features. These resulted in the PDDL2.1 language (Fox & Long, 2003). The extensive changes to PDDL2.1 and the ambitious objectives of the competition help to account for the fact that fewer people participated in 2002 than in 2000. Once again, the real testing and gathering of data took place over the two months prior to the conference. Although initial results were presented at the conference, no detailed analysis took place at the conference itself. The rest of this paper examines the objectives of the third competition, the results and some future challenges for the series.

## 3. Overview: The Third International Planning Competition

As organisers of the Third International Planning Competition we chose to address several new challenges that we believe are important to the ambitions of the planning community: the management of problems with metric constraints on numerically valued variables, temporal planning (including managing concurrency and scheduling of activities) and the construction of plans subject to specified optimisation criteria other than a simple count of the numbers of steps.

Setting these goals has obvious implications for the potential competitors in terms of extended expressive power and additional problem-solving required to manage the extensions. In order to control the extent to which competitors would be required to handle all of these extensions successfully, we constructed variants, or *levels*, of all of the domains used in the competition. For most of the domains we included a STRIPS level, a NUMERIC level (using STRIPS and metric variables only), a SIMPLETIME level in which actions have duration but there are no other metric values in the domain and a full temporal level, TIME, using durative actions with durations determined by the context of their usage (so duration depends on the parameter values and not only on the action name). The SIMPLETIME and TIME levels did not involve numeric resources other than time. To address this combination we introduced some additional domain variants, as we discuss below. Each of the four levels corresponds to a particular degree of expressive power of PDDL2.1, and there are different challenges posed by the versions of each domain at each level.

A secondary goal was to assess the relative effort of generating and encoding control rules for planners. Unfortunately, we failed to find any way to usefully quantify this effort. We discuss this question further in the following sections.

### 3.1 Problem Domains

The problem domains selected for the competitions have been, or have become, benchmark domains used by much of the community for empirical evaluation. The domains that have been used have often been chosen to probe some specific detail of performance. This has sometimes meant that the domains are not representative of general features of planning and are inappropriate for use in more widespread testing. A description of the domains used in all of the competitions so far can be found in Appendix A.

In the third competition, eight families of domains were used, broadly divided into transportation domains (Depots, DriverLog and ZenoTravel), applications-inspired domains (Rovers and Satellite) and a small collection of others (Settlers, FreeCell and UM-Translog-2).

We briefly summarise the collection here and describe them in more detail in Appendix A.

- **Depots** This domain combines the transportation style problem of Logistics with the well-known Blocks domain. The Logistics domain exhibits a high degree of parallelism, since separate vehicles can be utilised concurrently. In contrast, the Blocks domain is characterised by significant goal interaction. Our intention in doing this was to discover whether the successes of planners in each of these domains separately could be brought together when the problems were combined.

- **DriverLog** This problem involves transportation, but with the twist that the vehicles must be supplied with a driver before they can move.

- **Zeno-travel** Another transportation problem, inspired by a domain used in testing the ZENO planner developed by Pemberthy and Weld (1994), in which people must embark onto planes, fly between locations and then debark, with planes consuming fuel at different rates according to speed of travel.

- **Satellite** This domain was inspired by the problem of scheduling satellite observations. The problems involve satellites collecting and storing data using different instruments to observe a selection of targets.

- **Rovers** This domain was motivated by the 2003 Mars Exploration Rover (MER) missions and the planned 2009 Mars Science Laboratory (MSL) mission. The objective is to use a collection of mobile rovers to traverse between waypoints on the planet, carrying out a variety of data-collection missions and transmitting data back to a lander. The problem includes constraints on the visibility of the lander from various locations and on the ability of individual rovers to traverse between particular pairs of waypoints.

- **Settlers** This domain revolves around the management of resources, measured using metric valued variables. Products must be manufactured from raw materials and used in the manufacture or transportation of further materials. New raw materials can be generated by mining or gathering. The objective is to construct a variety of structures at various specified locations.

- **UM-Translog-2** This domain is a PDDL2.1 encoding of a new variant of the UM-Translog (Wu & Nau, 2002) domain. This was generated for us by Dan Wu of the University of Maryland. This is essentially a transportation domain, but one that is significantly more complex than previous transportation benchmarks. In fact, this domain was introduced late in the competition and very little data was collected. It is therefore not discussed further in this paper.

We also reused the Freecell domain from the second competition. This domain presented a serious challenge to participants in 2000 and we were interested to see whether planning technology had surpassed this challenge in the intervening two years. Although the domain produced some interesting data we did not attempt to precisely measure the extent to which the 2002 performance surpassed that of 2000.

Each domain (other than Settlers, Freecell and UM-Translog-2) was presented to the competitors for at least the four different levels previously identified: STRIPS, NUMERIC SIMPLETIME and TIME. The problems presented at each of these levels comprised distinct *tracks* and the competitors were able to choose in which tracks they wished to compete. In addition to the four main tracks we also included two additional tracks, intended to explore particular ideas. These tracks did not necessitate the use of additional expressive power but simply allowed existing expressiveness to be combined to produce interesting planning challenges. For example, the HARDNUMERIC track consisted of problems from the Satellite domain that had very few logical goals. Plans were evaluated by a metric based on amount of data recorded rather than by determining whether a specified logical goal had been achieved. The challenge was for planners to respond to the plan metric and include actions that would acquire data. The COMPLEX track consisted of problems that combined temporal and numeric features. The challenge was to reason about resource consumption in parallel with managing temporal constraints. In total, we defveloped 26 domains, with 20 problem instances in each domain (a few, unintentially, ended up with 16 or 22 instances). In most domains there were an additional 20 instances of large problems intended for the hand-coded planners. In total there were nearly 1000 problem instances to be solved, of which about half were intended primarily for the fully-automated planners.

## 3.2 The Competitors

The population of competing planning systems has changed over the three competitions. Few systems have competed more than once and no system has appeared in all three competitions. In part, this is a reflection of the speed of development of planning systems, revealing the extent to which the technology of 1998 has been surpassed by that of 2002. It is also a reflection of the growing interest in the series, which has encouraged more competitors to come forward, and of the work involved in taking part which has discouraged previous competitors from repeating their effort. Entering the competition involves more than generating interesting and efficient approaches to solving the planning problem: it demands the ability to construct a system that can respond robustly and reliably when confronted with previously unseen challenges, including domains that were not amongst those used in development of the system. It must be sufficiently well-engineered that its performance is not dominated by poor programming rather than by the real algorithmic effort of solving the problem (careless choice of a data structure can undermine the opportunity to show off a clever new planning algorithm). For systems that use hand-coded rules to guide the planning system there is an additional demand: the time required to read and understand domains sufficiently to construct sensible control knowledge and to then encode and test the knowledge to achieve good performance. The time-table for the testing was relatively compressed (the entire problem suite was generated and delivered to the competitors over a two month period and testing was carried out remotely on a single machine), so those using hand-coded controls were forced to find time to analyse the domains, hypothesise and test rules in intense sessions.

Details of the competing systems can be found in Appendix B. To summarise, many of the fully-automated planners use relaxed plan heuristics to guide a heuristic search. LPG (Gerevini, Saetti, & Serina, 2003) uses local search to improve candidate plan structures formed in a Graphplan-style plan graph. Several planners (MIPS, FF and SAPA) also extend the use of relaxed plan heuristics to include numeric features. VHPOP is a partial-order planner, demonstrating that partial-order planning can be competitive with the more recently fashionable heuristic forward state-space search planners. For various reasons, several planners generated only small collections of results, which are disregarded in the analysis.

A few competitors used more than one version of their planner, or more than one parameter setting. We did not attempt to enforce the use of unique versions, but left it to competitors to ensure that we were informed where variations were used. Multiple versions of FF, MIPS and LPG were used. FF submitted almost all data for a version optimised for speed performance. A small subset of data was submitted for a version optimised for quality, showing that there are alternative criteria by which performance can be evaluated. In all of the analyses we report we have used the data generated by FF optimised for speed exclusively. MIPS also offered data in two variants, using slightly different parameter settings. In our analyses we use data for one variant (MIPS) exclusively, except in the case of the Satellite HARDNUMERIC problems in which we used the data from the other variant (MIPS.plain). LPG submitted data in three versions: one based on earliest plan produced, one based on best plan produced over a longer time span and a third which represented a compromise between speed and quality. In fact, since all of the results for the version optimised for quality were generated within a few minutes at most, we chose to use this data exclusively in the analyses that follow. This should be borne in mind when reviewing the results for comparative speed performance of the planners. The performance of LPG relies on certain parameter settings. In most cases, the parameters used for LPG were automatically set, but in a few cases some parameters were set by hand. In their paper, appearing in this issue, Gerevini, Saetti and Serena (2003) give new results of an experiment testing their planner with all parameters set automatically. In general they observe no significant difference in the performance of LPG with respect to the data provided for the competition.

The three hand-coded planners that competed represent two alternative approaches to planning: forward state-space search using control rules to prune bad choices and promote good choices, and hierarchical task network (HTN) planning.

There were 508 problems available to the fully-automated planners and MIPS was the only planner to attempt all of them. There were 904 problems available to the hand-coded planners and SHOP2 was the only planner to attempt all of these, solving almost all of them and solving the most problems overall. TLPLAN and TALPLANNER were the only planners that solved all problems attempted. Not all planners attempted all problems they were equipped to handle. In particular, SAPA did not attempt STRIPS problems, although it is capable of solving them.

Although the planning competitions have been a great source of data from the state-of-the-art planners and an encouragement and catalyst for progress in the field, they are also lively and exciting competition events. This means that there must be "winners". The choice of winners is left to the organisers and must be treated with caution. In summarising the results at the AIPS conference in Toulouse, we presented a table of results in the form shown in Figure 1, together with a selection of graphs showing relative performance of planners in terms of speed and quality of plans in several of the problem sets. It was hard to synthesise a comprehensive view in the short time between final data collection and the presentation (a matter of a couple of days, during the conference), so our initial assessments were based on a rather crude evaluation of the evidence. We chose to award a prize to LPG as the best performer of the fully-automated planners: it solved the most problems of all fully-automated planners, showing excellent performance in the time tracks. We also awarded a prize to MIPS which solved the second most problems and had the widest coverage of all the fully-automated planners. It is clear that FF produced exceptional performance in the numeric level problems and could well have been judged worthy of a prize for that. We chose to acknowledge the great difficulty for newcomers to the competition in building a system that is sufficiently robust to compete, especially when there is no team of programmers and researchers to support the effort. For this reason we awarded a prize to VHPOP as best newcomer, with its creditable performance in both STRIPS and SIMPLETIME problems.

| Planner | Solved | Attempted | Success Ratio | Tracks entered |
|---|---|---|---|---|
| FF | 237 (+70) | 284 (+76) | 83% (85%) | S, N, HN |
| LPG | 372 | 428 | 87% | S, N, HN, ST, T |
| MIPS | 331 | 508 | 65% | S, N, HN, ST, T, C |
| SHOP2 | 899 | 904 | 99% | S, N, HN, ST, T, C |
| Sapa | 80 | 122 | 66% | T, C |
| SemSyn | 11 | 144 | 8% | S, N |
| Simplanner | 91 | 122 | 75% | S |
| Stella | 50 | 102 | 49% | S |
| TALPlanner | 610 | 610 | 100% | S, ST, T |
| TLPlan | 894 | 894 | 100% | S, N, HN, ST, T, C |
| TP4 | 26 | 204 | 13% | N, ST, T, C |
| TPSYS | 14 | 120 | 12% | ST, T |
| VHPOP | 122 | 224 | 54% | S, ST |

Figure 1: Table showing problems attempted and solved by each of the planners in the third IPC. Tracks are S: STRIPS, N: NUMERIC, HN: HARDNUMERIC, ST: SIMPLETIME, T: TIME and C: COMPLEX. Note that FF attempted 76 additional problems intended for the handcoded planners and solved 70 of them successfully. IxTeT solved 9 problems with plans accepted by the validator and attempted a further 10 problems producing plans that could not be validated due to differences between plan syntax used by IxTeT and defined in PDDL2.1.

Turning to the hand-coded planners, we awarded a prize for best performance to TLPLAN, which tackled almost all of the problems available, solved all of those it attempted and produced plans of very high quality with dramatic speed. We also rewarded SHOP2 for the fact that it attempted every problem and produced more plans than any other planner. The quality of its plans was consistently good and its performance highly competitive. TALPLANNER also performed outstandingly, often producing the highest quality plans and doing so tremendously efficiently, but its coverage was restricted to the STRIPS and TIME tracks. In selecting prize winners we chose to emphasise breadth of coverage, particularly in the new challenges of handling numeric and temporal planning problems. Competitions demand a degree of spectacle in the selection of winners, but the more measured evaluation of planners takes time. In this paper we present various analyses of the data collected during the competition and leave it to the community to judge the final rankings of the planners.

## 3.3 Additional Challenges

The HARDNUMERIC and COMPLEX problems used in the competition do not easily fit into the analysis conducted across the other results. These problems raise interesting special challenges for the planners. We now discuss some of these challenges, presenting data below. Each of these challenges was explored in only one or two problem sets, so generalisations about the performance of the planners based on the data collected are inappropriate. We do not, therefore, perform a statistical analysis of these data, but, instead, present the relevant data in simple graphical form.

### 3.3.1 The hardnumeric Satellite Problems

The HARDNUMERIC Satellite problem instances contained logical goals that are almost all trivial. For example, in most cases the problems involved simply ensuring that each of the satellites target a specific observation site at the end of the plan. However, the plan metric used to evaluate the plans was far more informative: the plans were evaluated according to the data collected by the satellites during their execution. Simply satisfying the explicit goals would generate a correct plan, but a worthless one in terms of the plan metric. Of the fully-automated planners, only MIPS and FF tackled these problems. TLPLAN and SHOP2 were the hand-coded planners that attempted these problems. It is instructive to compare the qualities of the plans produced by all four of these planners on this problem set. Figure 2 shows that the quality of the plans produced by the hand-coded planners is significantly higher than the quality of the plans generated by fully-automated planners. Indeed, FF generates plans that satisfy only the logical goals and minimise the plan size required to achieve that, so do not lead to any data collection. With some careful adjustments, MIPS has been able to generate plans that collect some data, but this is clearly a rather limited result. The closeness of the results generated by TLPLAN and SHOP2 suggest that they are both solving the data collection problem at a level close to optimal, or are applying similar heuristic approaches to the problem and generating similar locally optimal solutions. This domain very clearly highlights an advantage of hand-coded planners in their exploitation of the knowledge that their human domain-engineer can bring to bear.

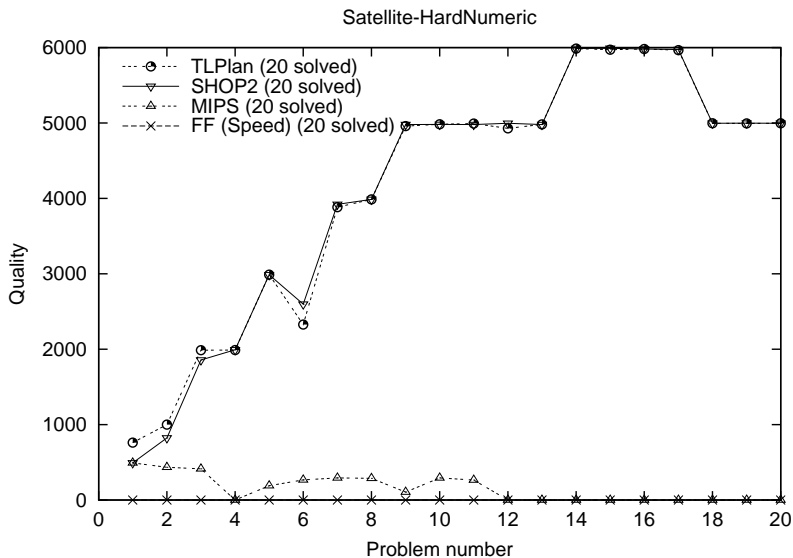

Figure 2: Plan quality for the Satellite HARDNUMERIC problems. High values are better quality plans: quality is the amount of data collected.

### 3.3.2 The Complex Satellite Domain

In addition to the HARDNUMERIC Satellite domain, a COMPLEX Satellite domain was also considered. The complexity in this domain arises from the fact that it combines temporal actions, with durations dependent on the parameters of the action, with the management of numerically measured resources (in this case, the data store available for acquired data). The problem has a quality that is similar to a knapsack problem, with data having to be packed into the limited stores of appropriate satellites. This is combined with the temporal optimisation problem, which involves ensuring that the satellites are efficiently deployed, moving between targets and capturing data according to their capabilities, their current aspects and their available store. As can be seen from Figure 3, the planners — fully-automated and hand-coded — produced plans of quite widely varying quality (lower values are better, here, since quality is measured by makespan). In general, TLPLAN produced the best quality plans, although LPG also produced some high quality plans for the smaller problems. As can be seen in problems 13, 16, 19 and 20, particularly, the fully-automated planners occasionally produced plans with quality diverging very significantly from optimal (we do not actually know what the optimal values are for these problems, but we can obviously consider the best value produced as an upper bound on the optimal value).

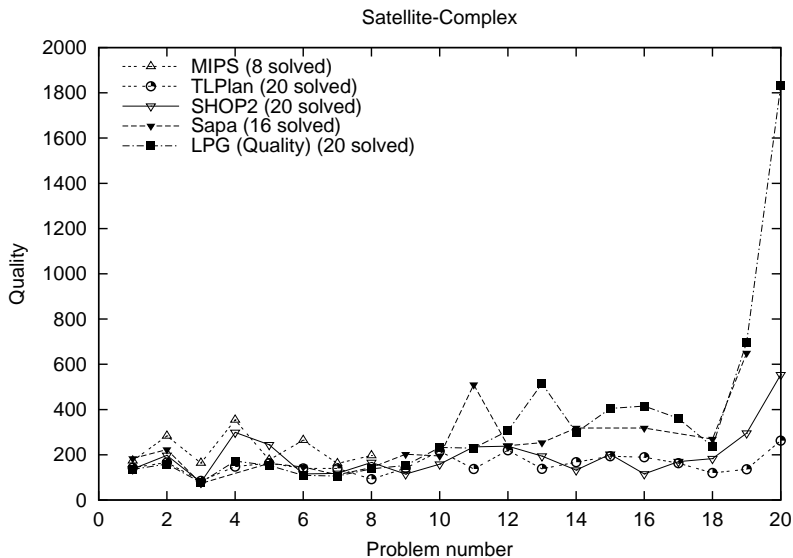

Figure 3: Plan quality for the Satellite COMPLEX problems. Low values represent better plans, since quality measures makespan, which was to be minimised.

### 3.3.3 The Settlers Domain

The Settlers domain is based on resource-management computer games, in which resources must be accumulated and used to construct new resource centres, with new specialised production and capabilities. An interesting problem that this domain highlights is that the PDDL2.1 family of languages does not offer a convenient way to name objects that are created during execution of a plan. In the version used for the competition, we overcame this problem by having a selection of names available at the outset that are not initially committed to any role. When an object is constructed, an unallocated name is used to name it, assigning to the name the properties of the new object and marking the name as used. An important problem that this approach creates for most current planners is that the initial grounding of actions involves creating all possible ways in which the names could be assigned to objects and then used in various roles within the plan. This leads to a significant explosion in the size of the domain encoding. It is interesting to observe that only 6 of the 20 problems were solved by any (fully-automated) planner, and that only FF solved more than one problem. It is clear that this domain remains a challenge to planning technology, but also to the future development of PDDL2.1 in which it will be necessary to review the way in which object construction is modelled.

## 4. Analysis of Competition Performance

One of the important roles of a competition is to identify the state of the art in the field, in terms of a variety of measures, and to attempt to answer broad questions about the landscape of the research area. Of course, a competition cannot have the last word about the state of the art. There might be state of the art planning systems that are prevented, for whatever reason, from taking part in the competition. However, the competition does provide some useful insights into the capabilities of modern planning systems.

Planning performance can be measured in terms of the speed of solution of the problems and the number and quality of the solutions found. Other metrics might also be identified. There are different planning architectures and heuristic evaluation functions as well as different kinds of domains and problems. The state of the art is represented by specific exemplars of these architectures and heuristic functions and it is interesting to explore the suitability of different architectures for use on different domains, the scaling behaviour of particular approaches, the comparative performance of different planning systems, etc. We address some of these issues in the following sections.

We perform three collections of analyses to ascertain comparative performance based on consideration of raw speed and quality data, the extent to which *domain* influenced planner behaviour and the scaling behaviour of the planners. The first collection is a comparison between the planners based on their raw competition performance. We analyse the data from the point of view of a consumer interested in ranking the planners according to speed and plan quality. These experiments are aimed at answering coarse level questions of the form: *which planner should I buy?* In asking questions such as this we are trying to arrive at a general basis for comparison. Of course, our investigation of this question is constrained by the metrics used in the competition. Furthermore, the trade-off one makes between time and quality depends on the context in which a planner might be deployed. Therefore, we cannot combine the judgements of relative speed and relative plan quality performance to determine unequivocally which planner to buy. We take as a basic assumption that potential users of domain-independent planning technology are interested primarily in broad coverage with good behaviour across a wide variety of domains, rather than in restricted coverage with spectacular behaviour in just a few domains. Our raw performance analyses are based mainly on the Wilcoxon rank-sum matched-pairs test (see Appendix C), as explained below. An advantage of this test is that it is non-parametric, meaning that it does not rely on assumptions about the shapes or properties of the underlying population distributions. It is also robust to outliers.

The second collection of experiments is concerned with identifying whether there were domains that were significantly easier (or harder). We perform these experiments at each of the levels of problems used in the competition to determine whether there is agreement amongst the planners about the difficulty of the problems. In part, this assists us in going on to explore scalability issues but it also allows us to determine whether the problem set presented at the competition contained interesting challenges. Our third collection of experiments compares the way in which planners scale on problem sets in which they agree about problem difficulty.

## 4.1 Raw Performance Analysis

We perform pairwise comparisons between planners to ascertain whether any consistent pattern can be identified in their relative speed and plan quality. We focus first on comparing the fully-automated planners and, separately, the hand-coded planners. We then perform an additional set of analyses to try to determine the raw performance benefit obtained from the use of hand-coded control knowledge. To do this, we perform the Wilcoxon test on pairs crossing the boundary between the fully-automated and hand-coded planner groupings. Where the conclusion is that the improvement obtained is significant, all we can say is that control rules yield an improvement in performance. We cannot account for the price, in terms of effort to encode the rules, that must be paid to obtain this improvement. The understanding of what is involved in writing useful control knowledge is still anecdotal and it remains an important challenge to the community to quantify this more precisely. One important consequence of the use of hand-crafted control knowledge is that to speak of "planner" performance blurs the distinction between the planning system itself and the control rules that have been produced to support its performance in each domain. Where a planner performs well it is impossible to separate the contributions from the planning system, the architecture of that system (and the extent that this contributes to the ease of expressing good control rules) and the sophistication of the control rules that have been used. We do not attempt to distinguish planner from control rules in the analysis that follows, but at least one competitor observed that results would have been significantly worse had there been less time to prepare, while, given more time, results could have been improved by concentrating on optimisation of plan metrics rather than simply on makespans. This observation helps to highlight the fact that, for planners exploiting hand-coded control knowledge, the competition format should be seen as a highly constrained basis for evaluation of performance.

To summarise, we now present the hypotheses we are exploring in this section:

> **Null Hypothesis:** There is no basis for any pairwise distinction between the performances of planners in terms of either time taken to plan or in quality (according to the specified problem metrics) of plans produced.

> **Alternative Hypothesis:** The planners can be partially ordered in terms of their time performances and, separately, their quality performances.

## 4.2 Analytic Framework

We perform pairwise comparisons, between planners in the fully-automated and hand-coded groups, on the problems in each of the main four problem levels used in the competition. We do not include analyses of the COMPLEX and HARDNUMERIC tracks because these resulted in too few data points for meaningful statistical conclusions to be drawn. We perform Wilcoxon rank-sum matched-pairs tests to identify whether the number of times one planner performed better than the other indicates a significant difference in performance between the two. Having performed pairwise comparisons between the performances in each of the tracks we use the results to construct partial orderings on the speed and quality performances of the planners in these tracks. We use 0.001 as the significance level because we wish to extrapolate from collections of pairwise comparisons to infer confidence, at the $p = 0.05$ level, in the transitive relationships between the planners. In the STRIPS track, which was the largest, we perform 15 pairwise comparisons so that a confidence level of $95^{1/15} = 0.003$ was required in the transitive picture. We use a confidence level of 0.001, resulting in a slightly conservative analysis.

We use sufficiently large samples that the T-distribution, used by the Wilcoxon test, is approximately normal. We therefore quote Z-values to indicate the significance in the differences between the mean performances of the paired planners. We do not compare all pairs of planners: in some cases the superiority of a planner over another is so clear from examination of the raw data that statistical testing is obviated.

The Wilcoxon test tells us when one planner performs consistently better than another and whether this consistency is statistically significant. It does *not* tell us how much better one planner is than another. It could be that there is very little difference in performance — perhaps the discrepancy can be accounted for by mere implementation differences. The consistency with which the difference occurs determines whether there is a statistically significant discrepancy between the performances of the planners. One planner can perform consistently better than another even though the other wins some of the time. For example, in a comparison between A and B, planner A might win more frequently than B because it is faster at solving a subset of problems in the set even though it is much slower at solving another subset of the problems. Provided the subset won by B is sufficiently large the greater number of wins due to A will not emerge as significant. For example, in a set of 10 problems ranked according to the differences in performance, if the first 7 problems in the ranking are won by A and the last 3 by B, A will obtain a score of 28 and B a score of 27. In this case no significant difference emerges *regardless* of the magnitude of the difference between A and B in the last three problems (see Appendix C). The rank-sum approach has the benefit of allowing large numbers of small wins to outweigh small numbers of large wins. This is desirable because a large win in a few problems is not an indication of overall better performance. We are interested in trying to measure the consistency of domain-independent behaviour and therefore in comparing the consistency of performance across domains given that planners perform differently within domains.

The size of the win need not indicate the complexity of the underlying problem so this does not allow us to make judgements about scalability. Essentially the test reveals consistent patterns in performance across a range of problems. We consider this to be of interest because knowing that one planner is significantly consistently better than another helps us make an objective judgement about which of the two planners is performing better across a varied problem set. If there is variability in the performances so that no consistent pattern emerges, it is very hard — perhaps impossible — to make this judgement objectively.

In a few cases, comparison using the Wilcoxon test does not lead to any statistically significant conclusions, but the proportion of wins by one of the planners is higher than is consistent with the null hypothesis of equal performance. This can be tested using a Z-test for a proportion (see Appendix C). Where this test yields a significant result we report it, as described below. This test is less informative than the Wilcoxon test as it does not take into account how the wins are distributed through the problem set.

In performing pairwise comparisons we must deal with cases where a planner did not solve a problem. We assign infinitely bad speed to the planner in these cases, ensuring that maximum possible benefit is given to planners solving problems, even very slowly. This methodology is valid because the Wilcoxon test is based on rank so the effect is simply to push the unsolved cases to the extreme of the ranking. In the case of quality we perform an initial collection of tests using infinitely bad quality for the unsolved problems. It is somewhat difficult to decide what it means to compare a true quality result with the infinitely bad quality assigned when a planner produced no solution. Our conclusion is that this comparison may not be informative enough, so in addition we perform the Wilcoxon test on just those cases where both planners produced a plan. We refer to these cases as *double hits*.

### 4.3 Results of Analysis

The results of performing the Wilcoxon tests, in order to compare speed performance between fully-automated planners, are shown in Figure 4. The results of similar tests to compare plan quality are presented in Figures 5 and 6. The double hits data are presented in Figure 6. The corresponding tests for the hand-coded planners are shown in Figures 7 and 8.

The tables have rows corresponding to the four problem levels for which the competition gathered sufficient data for analysis. These are: STRIPS, NUMERIC, SIMPLETIME and TIME. There are so many results for the fully-automated planners on STRIPS domains that they are split over two rows, creating five rows in these tables. In the comparisons of plan quality we report the STRIPS results using sequential plan length and concurrent plan length separately. The data in the rows are interpreted in the following way. Each cell, representing a pair of planners being compared, presents the Z-value and corresponding p-value identified from the Wilcoxon statistical table. The order of the planners' names in the title of the cell is significant: the first planner named is the one favoured by the comparison. Underneath the cell is an entry indicating the size of the sample used. The sample consists of all problems for which at least one of the planners being compared produced a solution: this results in there being different sample sizes for different comparisons. If the p-value is no greater than 0.001 then the difference in the mean performances obtained by the competing planners is statistically significant and it can be concluded that the planner in that column is significantly out-performing its competitor. If the p-value is greater than 0.001 the difference is not significant, in terms of the transitive view in which we are interested, and the null hypothesis that the planners are performing roughly equally cannot be rejected. We indicate the absence of significance at the $p < 0.001$ level by the use of a bold font.

The Wilcoxon test tells us when there is a significant difference in mean behaviour but it does not identify the planner producing the greater proportion of wins in cases where the mean behaviour is insignificantly different. Therefore, when the Wilcoxon tests reports that there is no significant difference between a pair of planners we also report the Z-value of the proportion (see Appendix C), if significant, to provide this missing information. Where

| | FF-LPG | LPG-MIPS | LPG-Sim | Sim-MIPS | MIPS-VHPOP | VHPOP-Stella |
|---|---|---|---|---|---|---|
| **Strips** | 6.2 | 5.3 | **1.9** | **1.9** (3.1) | 3.4 | **0.11** |
| | ⋆ | ⋆ | **0.06** | **0.06** (⋆) | ⋆ | **0.92** |
| | 120 | 118 | 118 | 114 | 98 | 59 |

| | Sim-Stella | LPG-VHPOP | FF-MIPS | FF-Sim | MIPS-Stella | Sim-VHPOP |
|---|---|---|---|---|---|---|
| **Strips** | 7.2 | 7 | 8.9 | 7.8 | 4.7 | 4.3 |
| | ⋆ | ⋆ | ⋆ | ⋆ | ⋆ | ⋆ |
| | 83 | 117 | 117 | 117 | 80 | 108 |

| | FF-LPG | LPG-MIPS | FF-MIPS | | | |
|---|---|---|---|---|---|---|
| **Numeric** | 3.5 | 3.8 | 8 | | | |
| | ⋆ | ⋆ | ⋆ | | | |
| | 93 | 86 | 85 | | | |

| | LPG-MIPS | MIPS-VHPOP | VHPOP-TP4 | LPG-TP4 | MIPS-TPSYS | TP4-TPSYS |
|---|---|---|---|---|---|---|
| **Simple Time** | 5 | **2** | 5.9 | 8.4 | 5.9 | **2.6** |
| | ⋆ | **0.04** | ⋆ | ⋆ | ⋆ | **< 0.01** |
| | 100 | 68 | 54 | 100 | 47 | 14 |

| | LPG-Sapa | MIPS-Sapa | LPG-MIPS | MIPS-TP4 | Sapa-TP4 | |
|---|---|---|---|---|---|---|
| **Time** | 3.3 | **0.72** | 3.4 | 5.1 | 5.2 | |
| | ⋆ | **0.67** | ⋆ | ⋆ | ⋆ | |
| | 95 | 72 | 96 | 36 | 38 | |

Figure 4: Table showing results of statistical tests for the comparison of speeds of planners. Bolded results are those that are *not* significant at p = 0.001 level. Each cell represents a pair of planners being compared. It presents the Z-value and corresponding p-value identified from the Wilcoxon statistical table. The order of the planners' names in the title of the cell is significant: the first planner named is the one favoured by the comparison. Underneath the cell is an entry indicating the size of the sample used. '⋆' indicates a result less than 0.001.

|  | LPG-FF | LPG-MIPS | LPG-Sim | MIPS-Sim | MIPS-VHPOP | VHPOP-Stella |
|---|---|---|---|---|---|---|
| **Strips (Seq)** | **0.21** | 6.9 | 7.6 | **0.38** | **1.3** | **2.1** (3.3) |
|  | **0.84** | ★ | ★ | **0.7** | **0.21** | **0.04** (★) |
|  | 120 | 118 | 118 | 114 | 98 | 59 |
|  | Sim-Stella | LPG-VHPOP | FF-MIPS | FF-Sim | MIPS-Stella | Sim-VHPOP |
| **Strips (Seq)** | 5.6 | 7.7 | 7.2 | 8.5 | 5 | **0.45** |
|  | ★ | ★ | ★ | ★ | ★ | **0.65** |
|  | 83 | 117 | 117 | 117 | 80 | 108 |
|  | LPG-FF | LPG-MIPS | LPG-Sim | MIPS-Sim | MIPS-VHPOP | VHPOP-Stella |
| **Strips (Conc)** | 6 | 5.3 | 8.6 | **2.4** (4.1) | **0.69** | **0.93** |
|  | ★ | ★ | ★ | **0.01** (★) | **0.49** | **0.35** |
|  | 120 | 118 | 118 | 114 | 98 | 59 |
|  | Sim-Stella | LPG-VHPOP | FF-MIPS | FF-Sim | MIPS-Stella | VHPOP-Sim |
| **Strips (Conc)** | **2.1** | 5.2 | **1.4** | 8.5 | 3.7 | **0.13** |
|  | **0.03** | ★ | **0.16** | ★ | ★ | **0.90** |
|  | 83 | 117 | 117 | 117 | 80 | 108 |
|  | LPG-FF | LPG-MIPS | FF-MIPS |  |  |  |
| **Numeric** | **3** (4.5) | 6.1 | 3.5 |  |  |  |
|  | **< 0.01** (★) | ★ | ★ |  |  |  |
|  | 93 | 86 | 85 |  |  |  |
|  | LPG-MIPS | MIPS-VHPOP | VHPOP-TP4 | LPG-TP4 | MIPS-TPSYS | TP4-TPSYS |
| **Simple Time** | 8.7 | **2.4** (3.5) | 5.4 | 8.6 | 5.6 | **2.7** |
|  | ★ | **0.01** (★) | ★ | ★ | ★ | **< 0.01** |
|  | 100 | 68 | 54 | 100 | 47 | 14 |
|  | LPG-Sapa | Sapa-MIPS | LPG-MIPS | LPG-TP4 | MIPS-TP4 | Sapa-TP4 |
| **Time** | 6.7 | **0.029** | 6.6 | 6.6 | 5 | 5.2 |
|  | ★ | **0.99** | ★ | ★ | ★ | ★ |
|  | 95 | 72 | 96 | 57 | 36 | 38 |

Figure 5: Table of results of statistical tests of comparisons of plan quality across problems solved by at least one planner in each pair. Bolded results are those that are *not* significant at the p = 0.001 level. '★' indicates a result less than 0.001.

| | LPG-FF | LPG-MIPS | LPG-Sim | MIPS-Sim | VHPOP-MIPS | VHPOP-Stella |
|---|---|---|---|---|---|---|
| **Strips (Seq)** | **0.24** | 4.3 | 5.9 | **1.4** | **2.9** | 4.7 |
| | **0.83** | ★ | ★ | **0.16** | **< 0.01** | ★ |
| | 114 | 85 | 90 | 63 | 56 | 39 |
| | Sim-Stella | LPG-VHPOP | FF-MIPS | FF-Sim | MIPS-Stella | VHPOP-Sim |
| **Strips (Seq)** | **1.8** | 3.6 | 4.6 | 7.1 | **3.1** | 5.2 |
| | **0.08** | ★ | ★ | ★ | **< 0.01** | ★ |
| | 49 | 68 | 86 | 91 | 44 | 51 |
| | LPG-FF | LPG-MIPS | LPG-Sim | MIPS-Sim | VHPOP-MIPS | VHPOP-Stella |
| **Strips (Conc)** | 6.5 | **1.5** | 7.5 | 6.3 | 3.9 | **2.7** |
| | ★ | **0.14** | ★ | ★ | ★ | **< 0.01** |
| | 114 | 85 | 90 | 63 | 56 | 39 |
| | Stella-Sim | VHPOP-LPG | MIPS-FF | FF-Sim | MIPS-Stella | VHPOP-Sim |
| **Strips (Conc)** | 6 | **3** | 4.8 | 7.1 | **0.24** | 6.1 |
| | ★ | **< 0.01** | ★ | ★ | **0.83** | ★ |
| | 49 | 68 | 86 | 91 | 44 | 51 |
| | LPG-FF | LPG-MIPS | MIPS-FF | | | |
| **Numeric** | 3.8 | 3.2 | 4.2 | | | |
| | ★ | ★ | ★ | | | |
| | 69 | 46 | 50 | | | |
| | LPG-MIPS | MIPS-VHPOP | TP4-VHPOP | LPG-TP4 | TPSYS-MIPS | TP4-TPSYS |
| **Simple Time** | 6.6 | **2.9 (3.8)** | 3.4 | **1.3** | **0.61** | **1.8** |
| | ★ | **< 0.01 (★)** | ★ | **0.19** | **0.54** | **0.07** |
| | 58 | 44 | 15 | 15 | 10 | 10 |
| | LPG-Sapa | MIPS-Sapa | LPG-MIPS | LPG-TP4 | TP4-MIPS | TP4-Sapa |
| **Time** | 4.7 | **1.6** | 4.2 | **1.9** | **1.1** | **1.3** |
| | ★ | **0.09** | ★ | **0.06** | **0.27** | **0.19** |
| | 62 | 50 | 55 | 5 | 5 | 5 |

Figure 6: Table showing results of statistical tests of comparisons between quality of plans from pairs of planners considering only problems solved by both planners. Each cell represents a pair of planners being compared. It presents the Z-value and corresponding p-value identified from the Wilcoxon statistical table. The order of the planners' names in the title of the cell is significant: the first planner named is the one favoured by the comparison. Underneath the cell is an entry indicating the size of the sample used. '★' indicates a result less than 0.001.

| | Small Problems | | | Large Problems | | |
|---|---|---|---|---|---|---|
| | TL-TAL | TAL-SHOP2 | TL-SHOP2 | TAL-TL | TAL-SHOP2 | TL-SHOP2 |
| Strips | 6.8 | **0.028** | 7.2 | 5.6 | 8.5 | 8.2 |
| | ⋆ | **0.99** | ⋆ | ⋆ | ⋆ | ⋆ |
| | 102 | 102 | 102 | 98 | 98 | 98 |
| | TL-SHOP2 | | | TL-SHOP2 | | |
| Numeric | 6.5 | | | 7.9 | | |
| | ⋆ | | | ⋆ | | |
| | 102 | | | 98 | | |
| | TL-TAL | SHOP2-TAL | TL-SHOP2 | TL-TAL | TAL-SHOP2 | TL-SHOP2 |
| Simple Time | 8.7 | **0.61** | 7 | **0.77** | 6.4 | 8.4 |
| | ⋆ | **0.34** | ⋆ | **0.44** | ⋆ | ⋆ |
| | 102 | 102 | 102 | 98 | 98 | 98 |
| | TL-TAL | TAL-SHOP2 | TL-SHOP2 | TL-TAL | TAL-SHOP2 | TL-SHOP2 |
| Time | 8.8 | **0.2** | 7.8 | 3.1 | 7.3 | 8 |
| | ⋆ | **0.84** | ⋆ | ⋆ | ⋆ | ⋆ |
| | 102 | 102 | 102 | 98 | 98 | 98 |

Figure 7: Table showing results of statistical tests for the comparison of speeds of hand-coded planners. '⋆' indicates a result less than 0.001.

| | Small Problems | | | Large Problems | | |
|---|---|---|---|---|---|---|
| | TL-TAL | TAL-SHOP2 | TL-SHOP2 | TAL-TL | TAL-SHOP2 | TL-SHOP2 |
| Strips (Seq) | **2.3** | **2** | 5.3 | **0.89** | 4.4 | 3.6 |
| | **0.01** | **0.04** | ⋆ | **0.38** | ⋆ | ⋆ |
| | 102 | 102 | 102 | 98 | 98 | 98 |
| | TAL-TL | TL-SHOP2 | SHOP2-TL | TAL-TL | TAL-SHOP2 | SHOP2-TL |
| Strips (Conc) | 8.6 | 4.3 | 7 | 8.6 | 7 | 7.7 |
| | ⋆ | ⋆ | ⋆ | ⋆ | ⋆ | ⋆ |
| | 102 | 102 | 102 | 98 | 98 | 98 |
| | TL-SHOP2 | | | TL-SHOP2 | | |
| Numeric | **0.18** | | | **0.15** | | |
| | **0.86** | | | **0.88** | | |
| | 102 | | | 98 | | |
| | TL-TAL | TAL-SHOP2 | TL-SHOP2 | TAL-TL | TAL-SHOP2 | TL-SHOP2 |
| Simple Time | **1** | 4.5 | 5.3 | **0.76** | 5.5 | 5.4 |
| | **0.32** | ⋆ | ⋆ | **0.44** | ⋆ | ⋆ |
| | 102 | 102 | 102 | 98 | 98 | 98 |
| | TL-TAL | SHOP2-TAL | TL-SHOP2 | TL-TAL | TAL-SHOP2 | TL-SHOP2 |
| Time | 4 | **0.26** | 3.9 | **2.3** | **0.54** | 5.7 |
| | ⋆ | **0.80** | ⋆ | **0.01** | **0.58** | ⋆ |
| | 102 | 102 | 102 | 98 | 98 | 98 |

Figure 8: Table showing results of statistical tests on comparative quality of plans produced by hand-coded planners. This table shows results for problems solved by at least one of the planners — results restricted to problems solved by both are insignificantly different, since the hand-coded planners all solved almost all problems attempted. '⋆' indicates a result less than 0.001.

we do this the Z-value of the proportion, and its p-value, appear in brackets following the Wilcoxon result.

## 4.4 Interpretation

Our results show that the null hypothesis can be rejected. Therefore, we adopt the alternative hypothesis and here discuss the resulting partial orders inferred from the data.

The data presented in Figures 4 to 8 can be interpreted in terms of partial orderings on the speed and quality performances of the fully-automated and hand-coded planners at each of the four problem levels. This can be done, at each level, simply by putting an ordering between pairs of planners from A to B when the Wilcoxon value for that pair is reported in the sub-column associated with A and is significant at the 0.001 level. The results are shown in Figures 9 to 12. In each of these figures sub-graphs associated with each of the four problem levels are identified. The presence of an arrow in a graph indicates that a statistically significant ordering exists. The absence of an arrow from A to B indicates that no statistically significant relationship between A and B was found at the corresponding problem level and therefore that no transitive ordering can depend on such a relationship.

### 4.4.1 Partial orderings based on speed

Figure 9 describes the speed comparisons that can be made between the fully-automated planners according to the Wilcoxon test. It can be observed that FF is significantly consistently faster than the other fully-automated planners at the STRIPS and NUMERIC levels (the significance of each of the arrows in the figure is sufficient to support transitive reasoning). Indeed, at the STRIPS and NUMERIC levels there is an interesting linear ordering between FF, LPG, MIPS and VHPOP (three of which were the prize-winners amongst the fully-automated planners) which is maintained between LPG, MIPS and VHPOP at the SIMPLETIME level. Despite the observation that SIMPLANNER was faster, in a significant proportion of the STRIPS problems than MIPS, there is no significant Wilcoxon relationship between them so that the four prize-winners comprise a spine of performance around which the other planners competing at these levels are clustered. The relationship breaks down at the TIME level because only LPG, MIPS, SAPA and TP4 participated. In this data set it can be seen that MIPS and SAPA are indistinguishable, with respect to the Wilcoxon test, but LPG is significantly consistently faster than both.

For comparing the hand-coded planners the competition used a collection of small problems and a collection of large problems at each problem level. The large problems were beyond the capabilities of any of the fully-automated planners. Interestingly, the hand-coded planners behaved differently in the small and large problem collections. This is most marked in Figure 10, at the STRIPS level, where the performances of TLPLAN and TALPLANNER are inverted in the small and large problem sets. In the small SIMPLETIME and TIME problems TLPLAN is consistently faster than either TALPLANNER or SHOP2, with TALPLANNER and SHOP2 being statistically indistinguishable in these data sets. TLPLAN is also consistently faster than TALPLANNER, which is in turn consistently faster than SHOP2, in the large TIME problems.

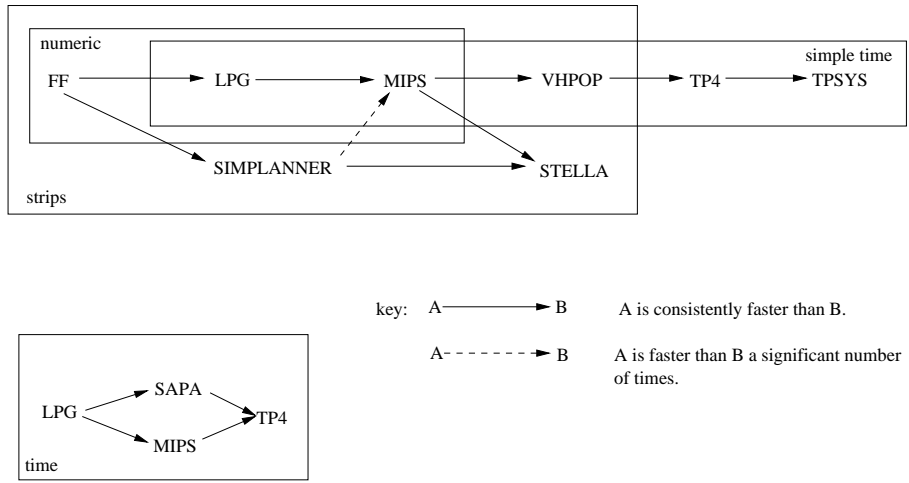

Figure 9: A partial order on the fully-automated planners in terms of their relative speed performances.

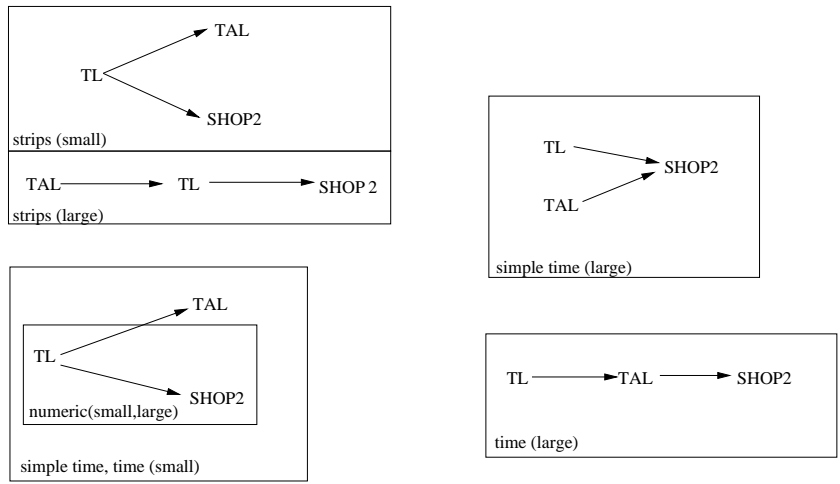

Figure 10: A partial order on the hand-coded planners in terms of their relative speed performances.

### 4.4.2 PARTIAL ORDERINGS BASED ON QUALITY

The construction of the partial order on quality performance at the STRIPS level for the fully-automated planners is shown in Figure 11. To interpret the figures depicting quality performance it should be noted that, at all problem levels except STRIPS, specific quality metrics were provided and plan quality must be measured in terms of these metrics. At the STRIPS level no such metrics were provided and quality is a measure of plan length — either sequential or concurrent. In the figures we have labelled the arrows in the STRIPS graphs to denote whether the relationship exists in terms of sequential or concurrent ordering, or both. Where the ordering is for both sequential and concurrent quality the arrow is left unlabelled. It can be observed that two planners might be ordered in one direction for sequential length and in the other for concurrent length.

As indicated above, comparison of quality performance is made difficult if one of the two planners being compared solved many more problems than the other (this problem only arises for the fully-automated planners because the hand-coded planners failed to solve so few problems that the proportion of unsolved problems did not affect our tests). Using an infinite quality measure for unsolved problems the Wilcoxon test concludes that the planner solving the most problems has overall better quality — in other words, if one is interested in overall solution quality one should choose the planner that solves the most problems even if, in some cases, it produces worse quality plans than its competitor. However, we also want to understand the relationship between the two planners in the double hits case. We notice that consideration of just these problems sometimes inverts the relationship detected by the first test. For example, in Figure 11 it can be observed that, at the SIMPLETIME level, VHPOP consistently produced better quality plans than TP4 across the whole problem set but, when only double hits are considered, TP4 produced consistently better plans than VHPOP. This suggests that TP4 is solving problems with higher quality solutions, but that the price it pays in search to find these solutions is so high that it solves a much smaller set of problems than other planners, such as VHPOP. We depict these results using dotted arrows in the graphs. Finally, it can arise that the Wilcoxon test detects no significant relationship between two planners, but that the difference in the proportion of problems solved by the two planners is significant. We indicate the resulting weaker relationship using a dashed arrow.

Figure 11 shows that LPG emerges as the fully-automated planner consistently producing the highest quality plans at all problem levels. The relationship between FF and MIPS is more complex because, whilst FF produced plans of better sequential quality than MIPS, MIPS produced better quality concurrent plans than FF when considering only double hits. The reason for this apparent discrepancy is that MIPS post-processes sequential plans into partially ordered plans exploiting some of the available concurrency in the problem, which FF does not exploit. However, it fails to solve a larger subset of problems than FF, giving FF the advantage in quality overall.

In the STRIPS problems SIMPLANNER solves more problems than STELLA and hence is seen to be performing at a consistently higher sequential-plan quality level. When double hits are considered STELLA outperforms SIMPLANNER in concurrent-plan quality. Also, when double hits are considered, it can be seen that VHPOP consistently outperforms STELLA for sequential-plan quality and SIMPLANNER in all cases. Interestingly, VHPOP and STELLA have no Wilcoxon or proportion test relationship when all problems are considered.

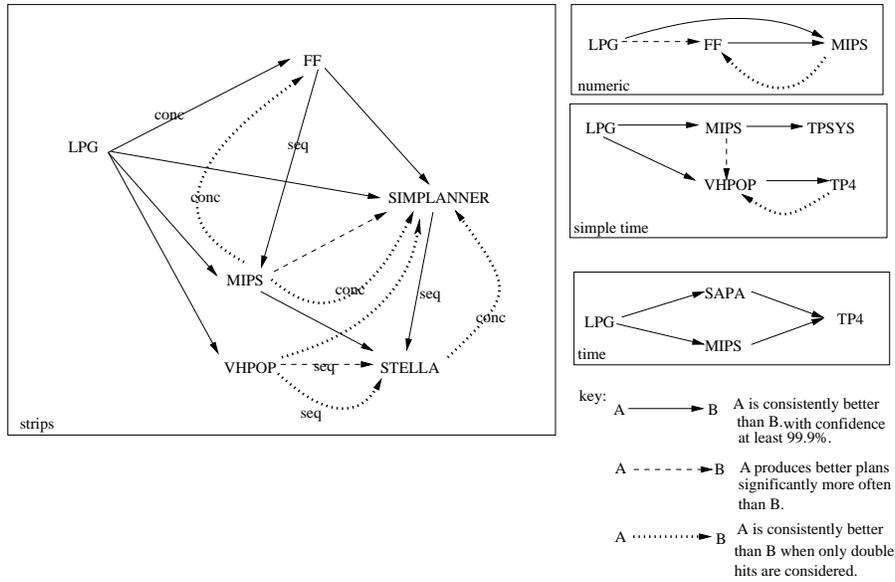

Figure 11: The fully-automated planners depicted in terms of their relative quality performances.

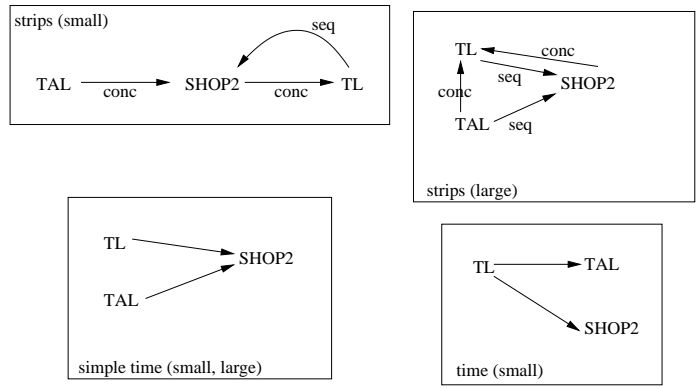

Figure 12: The hand-coded planners depicted in terms of their relative quality performances.

MIPS outperforms VHPOP and TPSYS in the SIMPLETIME problems, with VHPOP consistently better than TP4. When only double hits are considered TP4 outperforms VHPOP demonstrating that TP4 produces better quality solutions for those problems that it solves. Given the available data it seems that TP4 and TPSYS are not performing significantly differently, but it may be that the data set is too small to draw this conclusion with confidence.

In the TIME data set there is no significant consistent pattern in the relative performances of SAPA and MIPS. LPG consistently produces better quality plans.

As with the fully-automated planners we find that the speed comparisons that can be observed do not hold when the planners are compared in terms of quality. Figure 12 shows the quality comparisons we performed on the three competing hand-coded planners. It shows that, in the small STRIPS problems TLPLAN consistently outperformed SHOP2 in terms of sequential plan quality. In the small problems TALPLANNER produces shorter makespan plans than both TLPLAN and SHOP2 and SHOP2 produces shorter makespan plans than TLPLAN. TLPLAN produces sequential STRIPS plans and does not use any post-processing step to introduce parallelism. As a result it is certain to be outperformed in a makespan comparison with planners capable of producing parallel plans.

No significant relationships emerged in the NUMERIC problems. TALPLANNER did not compete in the NUMERIC problems. No significant Wilcoxon result was established.

In the SIMPLETIME problems (both small and large) the quality performances of TLPLAN and TALPLANNER are indistinguishable, and both are consistently better than SHOP2. In the TIME problems TLPLAN emerges as consistently better than TALPLANNER and SHOP2.

### 4.4.3 CROSS-BOUNDARY PARTIAL ORDERINGS

We performed a final collection of comparisons to try to better understand what advantages can be obtained from the use of hand-coded rather than fully-automated planners, in terms of speed and quality. We compare the best-performing fully-automated planner with the best-performing hand-coded planner in both categories: FF with TLPLAN for speed, at all levels, and LPG with TALPLANNER at the STRIPS level, SHOP2 at the numeric level and with TLPLAN at the remaining problem levels, for quality.

The tables in Figures 14 and 15 show the results of the tests. Figure 13 summarises the conclusions. It can be observed that TLPLAN is consistently faster than FF at all problem levels in which they both participated, demonstrating that the control knowledge being exploited by TLPLAN is giving it a real speed advantage. It remains to be seen exactly why this should be the case, given that for several STRIPS domains the control knowledge that is usually described as having been encoded appears to prune no additional states over those already pruned when an FF-style heuristic measure is used. The reason for this added value is an interesting question for the community to consider in trying to evaluate the advantages and disadvantages of the hand-coded approach.

It can also be observed that TALPLANNER produces consistently better concurrent plans than LPG at the STRIPS level. Again, this result needs to be explained by an in-depth analysis of the control information being exploited by TALPLANNER. At the SIMPLETIME level LPG produces plans that are consistently better quality than those of TLPLAN.

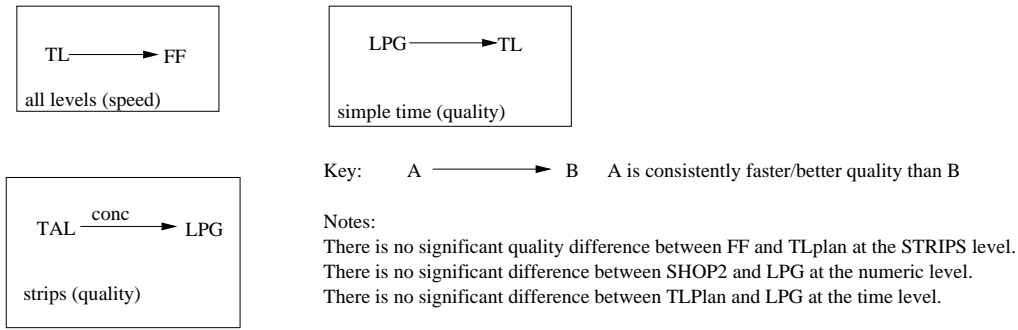

Figure 13: A comparison between the best of the fully-automated planners and the best of the hand-coded planners at each problem level.

| | TLPlan-FF | TAL-LPG | TAL-VHPOP |
|---|---|---|---|
| Strips | 6.7 | 7.8 | 8.5 |
| | < 0.001 | < 0.001 | < 0.001 |
| | 102 | 102 | 102 |
| Numeric | SHOP2-LPG | | |
| | 8.2 | | |
| | < 0.001 | | |
| | 102 | | |
| SimpleTime | TLPlan-LPG | | |
| | 8.6 | | |
| | < 0.001 | | |
| | 102 | | |
| Time | TLPlan-LPG | | |
| | 8.7 | | |
| | < 0.001 | | |
| | 102 | | |

Figure 14: Table of results for comparisons of fully-automated and hand-coded planners in terms of speed. Each cell represents a pair of planners being compared. It presents the Z-value and corresponding p-value identified from the Wilcoxon statistical table. The order of the planners' names in the title of the cell is significant: the first planner named is the one favoured by the comparison. Underneath the cell is an entry indicating the size of the sample used.

| | Problems solved by at least one | | | Problems solved by both | | |
|---|---|---|---|---|---|---|
| | TLPlan-FF | LPG-TAL | TAL-VHPOP | FF-TLPlan | LPG-TAL | VHPOP-TAL |
| STRIPS (Seq) | **0.57** | **1.8** (3.9) | **2.9** | **0.35** | **2.4** (4.2) | 4.9 |
| | **0.57** | **0.08** (< 0.001) | **< 0.01** | **0.72** | **0.01** (< 0.001) | < 0.001 |
| | 102 | 102 | 102 | 97 | 99 | 67 |
| | TLPlan-FF | TAL-LPG | TAL-VHPOP | FF-TLPlan | TAL-LPG | TAL-VHPOP |
| STRIPS (Conc) | **0.57** | 5.9 | 6.4 | **0.35** | 5.6 | **2.6** |
| | **0.57** | < 0.001 | < 0.001 | **0.72** | < 0.001 | **< 0.01** |
| | 102 | 102 | 102 | 97 | 99 | 67 |
| | SHOP2-LPG | | | LPG-SHOP2 | | |
| NUMERIC | **1.9** | | | **1.6** | | |
| | **0.06** | | | **0.11** | | |
| | 102 | | | 83 | | |
| | LPG-TLPlan | | | LPG-TLPlan | | |
| SIMPLE TIME | 3.9 | | | 4.3 | | |
| | < 0.001 | | | < 0.001 | | |
| | 102 | | | 100 | | |
| | TLPlan-LPG | | | LPG-TLPlan | | |
| TIME | **0.093** | | | **1.6** | | |
| | **0.92** | | | **0.11** | | |
| | 102 | | | 93 | | |

Figure 15: Table of results of comparisons of plan quality between fully-automated and hand-coded planners.

It is interesting to observe that hand-coding control information does not appear to lead to any consistent improvement in plan quality across the data sets. It does seem to lead to a speed advantage which must indicate that, in general, control rules provide a basis for more efficient pruning than weak general heuristic measures. The Wilcoxon test does not measure the extent of the speed advantage obtained, nor does it measure the extent of the quality advantage obtained from using a fully-automated planner in preference. These trade-offs need further close analysis, but it is interesting to see that there was not in fact a uniform advantage obtained by the hand-coded planners, at least on the smaller problems that formed the common foundation for testing. Of course, the development of hand-coded control knowledge can prioritise different aspects of the solutions generated and it is possible that further development of control rules might support the construction of more heavily optimised plans.

## 5. Tests for Magnitude

To complement the Wilcoxon tests we perform some additional analyses to identify whether, given two planners being compared, the *magnitude* of the difference in performance between the two planners is statistically significant. We perform paired t-tests (see Appendix C) using a subset of pairs of planners. We focus our attention on those pairs for which consistent significant differences were identified, because we consider it not to be meaningful to compare magnitude results for planners where no consistent domination is exhibited. We also restrict our attention to the planners that were, according to the Wilcoxon tests, the most impressive performers at each of the competition levels. We perform separate tests for speed and quality.

When investigating the magnitude of differences in performances it is not meaningful to include problems which were not solved by one of the planners being compared. Using infinite time or quality measures would result in a magnitude value that would grossly distort the true picture. For the magnitude tests we therefore consider only double hits. The price we pay for this is that we give undesirable emphasis to the smaller and easier problems since these are the ones most frequently solved by both planners. This should be borne in mind when interpreting the data.

The hypotheses being considered in this section are:

> **Null Hypothesis:** There is no consistent magnitude difference in the performances between planners.

> **Alternative Hypothesis:** Planners that demonstrate significant differences in consistency of performance also demonstrate corresponding magnitudes in the differences between their performances.

## 5.1 Analytic Framework

The t-tests are performed using the normalised performances of the two planners. They find the magnitude of difference between the performances of two planners, $p^1$ and $p^2$, on a collection of problem instances. For example, given a collection of $n$ problems, we find the pairs of results $p^1_{r_i}$ and $p^2_{r_i}$ obtained for instances $i = 1$ to $i = n$. In each case we normalise these values by dividing each of them by the mean of the pair. This process establishes a range of performances between 0 and 2, with 1 standing for equal performance. The t-test results in a t-value representing the difference in the mean normalised performances of the two planners. We perform 2-tailed tests at $p < 0.05$ because we are interested in the individual results rather than in extrapolated partial orderings based on magnitude. We want to consider the magnitude information as it is relevant to individual consistency results.

## 5.2 Results of Analysis

The tables in Figures 16 to 20 are organised as follows. Tables in Figures 16, 18 and 20 contain the speed results found for the fully-automated, hand-coded and mixed pairs respectively. Tables in Figures 17, 19 and 20 contain the quality results for the same three groups. In each table there is a row for each of the five competition levels (although empty rows have been omitted). The columns represent the pairs of planners being compared. In each cell five pieces of data are reported: the mean normalised performance for each planner; the t-value computed and the degrees of freedom used (which is derived from the number of double hits at that level) and the resulting p-value. A positive t-value means that the magnitude difference is in favour of the planner identified second in the cell. A negative t-value is in favour of the planner identified first. Where the resulting t-value indicates a difference in magnitude that is not significant at the p=0.05 level we use a bold font. In both speed and quality tests, an average performance smaller than 1 is favourable for a planner. The interpretation of the value is that it represents the average proportion of the mean performances of a pair of planners on each test set. Thus, an average performance of 0.66 for a planner (which will compare with an average performance of 1.34 for the other planner in the pair being considered) means that the first planner is, on average, twice as fast as the second.

| STRIPS | FF 0.4 | MIPS 1.21 | FF 0.22 | LPG 0.76 | LPG 1.28 | Sim 0.26 |
|---|---|---|---|---|---|---|
| | LPG 1.6 | LPG 0.79 | MIPS 1.78 | VHPOP 1.24 | Sim 0.72 | VHPOP 1.74 |
| | -12.26,113 | 2.69,84 | -21.30,85 | -2.82,67 | 3.73,89 | -15.19,50 |
| | < 0.001 | 0.007 | < 0.001 | 0.005 | < 0.001 | < 0.001 |
| NUMERIC | FF 0.23 | MIPS 0.8 | | | | |
| | LPG 1.77 | LPG 1.2 | | | | |
| | -16.04,68 | **-1.84,45** | | | | |
| | < 0.001 | **0.06** | | | | |
| SIMPLETIME | LPG 1.03 | MIPS 0.85 | | | | |
| | MIPS 0.97 | VHPOP 1.15 | | | | |
| | **0.35,57** | **-1.43,43** | | | | |
| | **0.73** | **0.15** | | | | |
| TIME | LPG 1.25 | LPG 1.16 | MIPS 0.77 | | | |
| | MIPS 0.75 | Sapa 0.84 | Sapa 1.23 | | | |
| | 2.64,54 | **1.81,61** | -3.41,49 | | | |
| | 0.008 | **0.07** | < 0.001 | | | |

Figure 16: Magnitude comparisons for fully-automated planners in terms of speed.

| STRIPS (seq) | MIPS 1.05 | FF 0.95 | LPG 0.98 | LPG 0.91 | Sim 1.08 |
|---|---|---|---|---|---|
| | LPG 0.95 | MIPS 1.05 | VHPOP 1.02 | Sim 1.09 | VHPOP 0.92 |
| | 5.61,84 | -5.39,85 | -2.73,67 | -7.21,89 | 6.01,50 |
| | < 0.001 | < 0.001 | 0.006 | < 0.001 | < 0.001 |
| STRIPS (conc) | FF 1.18 | FF 1.13 | LPG 0.7 | Sim 1.33 | |
| | LPG 0.82 | MIPS 0.87 | Sim 1.3 | VHPOP 0.67 | |
| | 9.70,113 | 6.81,85 | -12.95,89 | 14.37,50 | |
| | < 0.001 | < 0.001 | < 0.001 | < 0.001 | |
| NUMERIC | FF 1.17 | FF 1.01 | MIPS 1.2 | | |
| | LPG 0.83 | MIPS 0.99 | LPG 0.8 | | |
| | 6.00,68 | **0.14,49** | 4.25,45 | | |
| | < 0.001 | **0.89** | < 0.001 | | |
| SIMPLETIME | LPG 0.86 | MIPS 0.93 | | | |
| | MIPS 1.14 | VHPOP 1.07 | | | |
| | -7.98,57 | -3.97,43 | | | |
| | < 0.001 | < 0.001 | | | |
| TIME | LPG 0.9 | LPG 0.86 | | | |
| | MIPS 1.1 | Sapa 1.14 | | | |
| | -3.44,54 | -4.99,61 | | | |
| | < 0.001 | < 0.001 | | | |

Figure 17: Magnitude comparisons for fully-automated planners in terms of quality.

## 5.3 Interpretation

The results demonstrate that the null hypothesis can be rejected in almost all pairwise-comparisons of planners for which the Wilcoxon test shows a significant consistent performance difference. There are some cases in which the null hypothesis cannot be rejected, implying that the consistent performance difference in a pair of planners does not translate into a statistical significance in their mean relative performances.

| | Small Problems | | Large Problems | | |
|---|---|---|---|---|---|
| STRIPS | TLPlan 0.52 | TLPlan 0.61 | TLPlan 1.32 | TLPlan 0.39 | TAL 0.24 |
| | TAL 1.48 | SHOP2 1.39 | TAL 0.68 | SHOP2 1.61 | SHOP2 1.76 |
| | -12.98,101 | -8.32,101 | 7.74,97 | -17.41,97 | -26.04,97 |
| | < 0.001 | < 0.001 | < 0.001 | < 0.001 | < 0.001 |
| NUMERIC | TLPlan 0.7 | | TLPlan 0.31 | | |
| | SHOP2 1.3 | | SHOP2 1.69 | | |
| | -6.21,101 | | -19.22,92 | | |
| | < 0.001 | | < 0.001 | | |
| SIMPLETIME | TLPlan 0.43 | TLPlan 0.67 | TLPlan 0.32 | TAL 0.48 | |
| | TAL 1.57 | SHOP2 1.33 | SHOP2 1.68 | SHOP2 1.52 | |
| | -21.34,101 | -7.79,101 | -24.27,97 | -11.28,97 | |
| | < 0.001 | < 0.001 | < 0.001 | < 0.001 | |
| TIME | TLPlan 0.44 | TLPlan 0.59 | TLPlan 0.82 | TLPlan 0.31 | TAL 0.46 |
| | TAL 1.56 | SHOP2 1.41 | TAL 1.18 | SHOP2 1.69 | SHOP2 1.54 |
| | -25.00,101 | -10.14,101 | -5.75,97 | -23.75,95 | -12.81,95 |
| | < 0.001 | < 0.001 | < 0.001 | < 0.001 | < 0.001 |

Figure 18: Magnitude comparisons for hand-coded planners in terms of speed.

| | Small Problems | | | Large Problems | | |
|---|---|---|---|---|---|---|
| STRIPS (seq) | TLPlan 0.96 | | | TLPlan 0.96 | TAL 0.98 | |
| | SHOP2 1.04 | | | SHOP2 1.04 | SHOP2 1.02 | |
| | -5.48,101 | | | -3.82,97 | -2.13,97 | |
| | < 0.001 | | | < 0.001 | 0.033 | |
| STRIPS (conc) | TLPlan 1.38 | TLPlan 1.27 | TAL 0.88 | TLPlan 1.7 | TLPlan 1.5 | TAL 0.75 |
| | TAL 0.62 | SHOP2 0.73 | SHOP2 1.25 | TAL 0.3 | SHOP2 0.5 | SHOP2 1.25 |
| | 21.10,101 | 11.81,101 | -5.15,101 | 48.66,97 | 16.58,97 | -8.94,97 |
| | < 0.001 | < 0.001 | < 0.001 | < 0.001 | < 0.001 | < 0.001 |
| SIMPLETIME | TLPlan 0.89 | TAL 0.93 | | TLPlan 0.85 | TAL 0.87 | |
| | SHOP2 1.11 | SHOP2 1.07 | | SHOP2 1.15 | SHOP2 1.13 | |
| | -7.66,101 | -3.82,101 | | -6.85,97 | -4.80,97 | |
| | < 0.001 | < 0.001 | | < 0.001 | < 0.001 | |
| TIME | TLPlan 0.9 | TLPlan 0.94 | | TLPlan 0.88 | | |
| | TAL 1.1 | SHOP2 1.06 | | SHOP2 1.12 | | |
| | -4.39,101 | -4.08,101 | | -6.46,95 | | |
| | < 0.001 | < 0.001 | | < 0.001 | | |

Figure 19: Magnitude comparisons for hand-coded planners in terms of quality.

| | Speed | Quality | | | |
|---|---|---|---|---|---|
| | | Seq | | Conc | |
| STRIPS | TLPlan 0.55 | TLPlan 1 | TAL 1.04 | TAL 0.84 | TLPlan 1.23 |
| | FF 1.45 | FF 1 | LPG 0.96 | LPG 1.16 | LPG 0.77 |
| | -7.60,96 | **0.54,96** | 3.89,98 | -7.16,98 | 12.54,98 |
| | < 0.001 | **0.589** | < 0.001 | < 0.001 | < 0.001 |
| SIMPLETIME | LPG 1.9 | LPG 0.94 | | | |
| | TLPlan 0.1 | TLPlan 1.06 | | | |
| | 38.14,99 | -3.89,99 | | | |
| | < 0.001 | < 0.001 | | | |
| TIME | LPG 1.9 | | | | |
| | TLPlan 0.1 | | | | |
| | 26.85,92 | | | | |
| | < 0.001 | | | | |

Figure 20: Comparisons between fully-automated and hand-coded planners in terms of both speed and quality.

As shown in tables 16 to 20, a reassuring consistency emerged with the results of the Wilcoxon tests. That is: where significant consistency differences are identified between two planners using the Wilcoxon test, the t-test for magnitude generally reveals a significant magnitude difference as well.

## 6. Dependence of Performance on Domain

We consider it important to quantify the difficulty of the problems used in the competition because this provides the basis for a deeper understanding of relative planner performance. To explore this we investigate whether any of the domains that were used uniformly considered easy, or hard, amongst the fully-automated and hand-coded planners. We also investigate whether, as might be expected to be the case, the STRIPS problems were generally considered to be easier than the problems at the numeric and temporal levels. These two questions lead to two related investigations based on bootstrapping techniques. Our analyses show that different planners experienced different domains and levels as difficult, within both the fully-automated and the hand-coded categories.

In the first competition, in 1998, it was reported (Long, 2000) that no planner solved any problem with more than about 25,000 ground actions and that 10,000 ground actions marked a limit beyond which planner performance was markedly unreliable. A ground action is formed by replacing action schema parameters with objects of the correct types selected from those in a problem instance. Any static preconditions — preconditions whose truth can be ascertained entirely from the initial state — are used to filter the ground actions, so that only those that are plausibly applicable are counted. Relevant ground actions are found by applying a reachability analysis from the initial state and a regression analysis from the goals in order to identify the subset of ground actions that could actually play any useful role in the plan. It seems plausible that the number of ground actions could offer a guide to the difficulty of problems. In fact, a brief survey of the largest problems in the third competition, in Figure 21, reveals that action counts can vary widely across domains. It is encouraging to observe that the size of problems that can be solved with reasonable reliability, at least in some domains, has grown significantly, despite the fact that it is still typical for planners to ground the action set prior to planning. It is also of interest to observe that the size of problems measured by action counts is not a strong indication of difficulty — problems from the Rovers and Satellite domains were amongst those found harder by many of the planners, despite having small action counts.

To summarise, the hypotheses being explored in this section are:

> **Null Hypothesis:** The domains used in the competition were equally challenging to all planners at all levels.

> **Alternative Hypothesis:** Domain/level combinations can be distinguished in terms of relative difficulty posed to different planners.

For ease of comparison with the results presented in Sections 7 and 8 we observe that, in this section, we are specifically concerned with a cross-domain analysis and with whether the planners agreed on which of the domain/level combinations were hard.

| Domain | Largest Ground Action Count | Largest Relevant Action Count |
|---|---:|---:|
| Depots | 332,064 | 22,924 |
| DriverLog | 31,140 | 15,696 |
| ZenoTravel | 32,780 | 32,780 |
| Rovers | 7,364 | 3,976 |
| Satellite | 4,437 | 4,437 |
| FreeCell | 112,600 | 25,418 |
| Settlers | 5,884 | 4,503 |

Figure 21: Counts of ground action instances (generated using FF).

## 6.1 Analytic Framework

In order to explore the two questions, we used the planners to discover how hard the domains and levels were. For each planner, domain and problem level we plot the number of problem instances left to solve against time in milliseconds. This results in a curve, the area under which can be taken to be a measure of the difficulty experienced by the planner in solving problems in the given domain at the given problem level. In order to keep the area under the curve finite we use a cut-off time of thirty minutes. This extended cut-off time (fifteen minutes was used in the competition) results in a higher penalty being paid by a planner that fails to solve problems.

In the experiment used to address the first question, the null hypothesis is that the planner finds all problems at a specific level equally difficult across all domains. To test this we constructed, using a bootstrapping technique, ten thousand samples of twenty values from the collection of all timings obtained from domains at the appropriate level. The values were selected at random from the performances of planners competing in the domains, one value for each of a collection of randomly selected problems. For example, if problem one was chosen from DriverLog, problem two from Depots, problem three from Rovers, problem four from Depots, etc., then the value associated with problem one would be that produced for that problem by a planner selected at random from those that competed in DriverLog. Similarly, the value associated with problem two would be chosen from a planner that competed in Depots, and so on. For each collection of 20 values we plotted the number of problem instances left to solve against time, as above. This resulted in a sampling distribution of *level-specific* areas. Using these bootstrap samples we check whether the area calculated for the particular planner-domain-level combination lies at the extremes of this distribution, or not. If it lies in the first 2.5% of the distribution we reject the null hypothesis on the grounds that the planner found problems at that level, in that domain, to be significantly *easy*. If it lies in the top 2.5% of the distribution we reject the null hypothesis and conclude that those problems were significantly *hard* for that planner.

In testing the relative hardness of problem levels within a domain (the second question), we perform similar experiments in which, for each planner, the bootstrapped samples were obtained by sampling timings from all problem levels within all domains. This resulted in a new sampling distribution of the *level-independent* area statistic. The null hypothesis, that the domain/level combination is not an indicator of difficulty, is tested by seeing whether the areas computed for planner-domain-level combinations are extreme with respect to the new sampling distribution.

| | Level-dependent | | | | Level independent | | | |
|---|---|---|---|---|---|---|---|---|
| | STRIPS [6] | NUMERIC [3] | SIMPLETIME [3] | TIME [3] | STRIPS [6] | NUMERIC [3] | SIMPLETIME [3] | TIME [3] |
| Depots | 1/3 | 1/0 | 0/1 | 0/1 | 2/2 | 1/1 | 0/1 | 0/1 |
| DriverLog | 0/1 | 2/0 | 1/0 | 2/0 | 1/0 | 0/0 | 1/0 | 1/0 |
| ZenoTravel | 3/0 | 2/0 | 1/0 | 1/0 | 4/0 | 2/0 | 0/0 | 0/1 |
| Rovers | 2/1 | 0/1 | 0/1 | 1/1 | 3/1 | 0/2 | 0/1 | 0/1 |
| Satellite | 4/0 | 0/2 | 1/0 | 1/0 | 4/0 | 0/2 | 2/0 | 1/0 |
| FreeCell | 1/2 | – | – | – | 2/2 | – | – | – |
| Settlers | – | 0/2 | – | – | – | 0/2 | – | – |

Figure 22: Comparisons of performance between domains for fully-automated planners.

## 6.2 Results of Analysis

Figures 22 to 25 report the results of the two experiments described above. Figures 22 and 23 describe the level-specific and level-independent comparisons we made using the fully-automated planners and the hand-coded planners respectively. The table for the hand-coded planners is further divided into two parts: the first five rows correspond to the small problems, the latter five rows to the large problems. The performance of the hand-coded planners on the large problems was measured using bootstrapped samples taken from the large problem collection.

The tables are organised as follows: the rows correspond to domains, as labelled, and the columns to the levels considered and the number of planners used. The number of planners varies between columns because different planners participated at the different domain levels. For example, more planners participated at the STRIPS level than at any of the others. When planners produced too little data to justify statistical analysis they were not included in the tests. Thus, of the eleven fully-automated planners in the competition seven produced enough data for analysis in these experiments.

The cells of the tables contain two integer values separated by a diagonal. The value on the left of the diagonal indicates the number of planners that found the problems in the corresponding domain and level significantly easy. The value on the right indicates the number that found those problems significantly hard. Thus, it can be seen in Figure 22 that of the six fully-automated planners that participated at the STRIPS level of the Depots domain, one found the problems easy and three found them hard. For the other two planners the areas calculated using the method explained above were not found to be sufficiently extreme for rejection of the null hypothesis. Broadly speaking (we discuss the interpretation of the data in detail below) the four left-hand columns tell us whether the problems in a particular domain and level were easy or hard relative to other problems *at that level*; the four right-hand columns tell us whether they were easy or hard relative to all other problems. In addition, the rows allow us to compare domains for relative difficulty: for example, none of the planners found ZenoTravel to be hard at any level relative to other problems at the same level, whilst Depots and Rovers were found to be hard by at least one competitor at all levels.

| | Level-dependent | | | | Level-independent | | | |
|---|---|---|---|---|---|---|---|---|
| | STRIPS [3] | NUMERIC [2] | SIMPLETIME [3] | TIME [3] | STRIPS [3] | NUMERIC [2] | SIMPLETIME [3] | TIME [3] |
| Depots | 2/0 | 2/0 | 1/0 | 3/0 | 3/0 | 2/0 | 3/0 | 3/0 |
| DriverLog | 0/0 | 2/0 | 0/0 | 3/0 | 3/0 | 1/0 | 2/0 | 0/0 |
| ZenoTravel | 3/0 | 1/0 | 3/0 | 3/0 | 3/0 | 1/0 | 3/0 | 2/0 |
| Rovers | 0/2 | 0/1 | 0/3 | 0/2 | 1/0 | 0/1 | 0/1 | 0/2 |
| Satellite | 2/1 | 0/1 | 2/1 | 0/1 | 2/0 | 0/0 | 2/1 | 0/1 |
| Depots (large) | 2/0 | 2/0 | 2/0 | 3/0 | 2/0 | 1/0 | 2/0 | 2/0 |
| DriverLog (large) | 0/1 | 1/0 | 0/1 | 1/1 | 0/0 | 0/0 | 0/1 | 0/1 |
| ZenoTravel (large) | 3/0 | 1/0 | 3/0 | 3/0 | 3/0 | 1/0 | 3/0 | 3/0 |
| Rovers (large) | 1/1 | 0/0 | 1/1 | 1/0 | 2/0 | 0/0 | 2/0 | 0/0 |
| Satellite (large) | 1/0 | 0/1 | 2/0 | 0/2 | 1/0 | 0/1 | 2/0 | 0/1 |

Figure 23: Comparisons of performance between domains using hand-coded planners.

The level-independent tests are reported in exactly the same way in the right-hand halves of Figures 22 (for the fully-automated comparison) and 23 (for the hand-coded comparison). These tables tell us whether the problems in a particular domain and level are easy or hard relative to problems from other domains *irrespective of level.*

The data presented in Figures 24 and 25 show which planners found which domain-level combinations easy or hard as discussed with reference to the tables in Figures 22 and 23. This information might contribute to an understanding about which planning approaches are likely to be suited to what kinds of problems, although further analysis would be needed to pursue this question.

The tables are organised as follows. There is a row for each of the individual planners, indicating which domain-level combinations were found to be easy or hard for the corresponding planner. Associated with the categorization of a combination as easy or hard is the p-value indicating the statistical significance of this finding. We have presented only the findings that were significant at the 5% level. Because this is a two-tailed test (we had no a priori knowledge to help us to determine whether a problem would be easy or hard) the critical value at the easy end is 0.025. At the hard end the critical value is 0.975. Figure 24 shows our findings for the fully-automated planners. Figure 25 shows the same information with respect to the hand-coded planners.

## 6.3 Interpretation

The results allow us to reject the null hypothesis in some cases, but not in others. We are able to determine significant differences in the relative hardness of domains as determined by specific planners, but there is also evidence of lack of consistency between the judgements of different planners. For example, there are some domain/level combinations that are found hard by certain planners and not by others.

The tables in Figures 22 and 23 allow us to determine which domains presented the most interesting challenges to the planners participating in the competition. Although it is difficult to draw firm conclusions from data that is really only indicative, some interesting patterns do emerge. For example, the level-specific data in Figure 22 shows that none of the fully-automated planners found ZenoTravel problems, at any levels, to be significantly hard by comparison with problems drawn from other domains at the same level. The Satellite

| | | Easy | | Hard | |
|---|---|---|---|---|---|
| FF | Depots Numeric | 0.015 | Rovers Numeric | 1 |
| | Depots Strips | 0.012 | Settlers Numeric | 1 |
| | FreeCell Strips | 0.017 | | |
| | Rovers Strips | 0 | | |
| | Satellite HardNumeric | 0 | | |
| | Satellite Strips | 0.0007 | | |
| | ZenoTravel Numeric | 0.0026 | | |
| | ZenoTravel Strips | 0.0015 | | |
| LPG | Rovers Strips | 0.0007 | Satellite Numeric | 1 |
| | Satellite SimpleTime | 0.019 | ZenoTravel Time | 0.98 |
| | Satellite Strips | 0.0001 | | |
| MIPS | DriverLog HardNumeric | 0.0046 | Depots Numeric | 0.99 |
| | DriverLog SimpleTime | 0.0094 | Rovers Numeric | 0.98 |
| | DriverLog Strips | 0.0088 | Rovers Time | 0.98 |
| | DriverLog Time | 0.0093 | Satellite Complex | 0.98 |
| | FreeCell Strips | 0 | Satellite Numeric | 1 |
| | Satellite HardNumeric | 0 | Settlers Numeric | 1 |
| | ZenoTravel Numeric | 0.01 | | |
| | ZenoTravel Strips | 0.0021 | | |
| Sapa | Satellite Time | 0.0017 | Depots Time | 1 |
| Simplanner | Depots Strips | 0.0003 | Rovers Strips | 1 |
| | ZenoTravel Strips | 0.013 | | |
| Stella | Satellite Strips | 0.016 | Depots Strips | 1 |
| | ZenoTravel Strips | 0 | FreeCell Strips | 1 |
| VHPOP | Rovers Strips | 0 | Depots SimpleTime | 1 |
| | Satellite SimpleTime | 0.0006 | Depots Strips | 1 |
| | Satellite Strips | 0.0004 | FreeCell Strips | 1 |
| | | | Rovers SimpleTime | 0.98 |

Figure 24: Easy/hard boundaries for fully-automated planners.

|  | Easy | | Hard | |
|---|---|---|---|---|
| SHOP2 | Depots Numeric | 0.0031 | Rovers Numeric | 1 |
|  | Depots SimpleTime | 0.0057 | Rovers Time | 1 |
|  | Depots Strips | 0.0001 |  |  |
|  | Depots Time | 0.0088 |  |  |
|  | Depots HC SimpleTime | 0.0001 |  |  |
|  | Depots HC Strips | 0 |  |  |
|  | Depots HC Time | 0.0006 |  |  |
|  | DriverLog HardNumeric | 0.015 |  |  |
|  | DriverLog SimpleTime | 0.015 |  |  |
|  | DriverLog Strips | 0.019 |  |  |
|  | Satellite SimpleTime | 0 |  |  |
|  | Satellite Strips | 0 |  |  |
|  | Satellite HC SimpleTime | 0 |  |  |
|  | Satellite HC Strips | 0 |  |  |
|  | ZenoTravel Numeric | 0.0018 |  |  |
|  | ZenoTravel SimpleTime | 0.0003 |  |  |
|  | ZenoTravel Strips | 0.0001 |  |  |
|  | ZenoTravel Time | 0.0043 |  |  |
|  | ZenoTravel HC Numeric | 0.0001 |  |  |
|  | ZenoTravel HC SimpleTime | 0 |  |  |
|  | ZenoTravel HC Strips | 0 |  |  |
|  | ZenoTravel HC Time | 0.0001 |  |  |
| TALPlanner | Depots SimpleTime | 0 | DriverLog HC SimpleTime | 1 |
|  | Depots Strips | 0 | DriverLog HC Time | 1 |
|  | Depots Time | 0.013 | Rovers SimpleTime | 0.99 |
|  | Depots HC SimpleTime | 0.0029 | Rovers Time | 1 |
|  | Depots HC Strips | 0 | Satellite SimpleTime | 0.98 |
|  | DriverLog Strips | 0 | Satellite Time | 1 |
|  | Rovers Strips | 0.0026 | Satellite HC Time | 1 |
|  | Rovers HC SimpleTime | 0.0009 |  |  |
|  | Rovers HC Strips | 0 |  |  |
|  | ZenoTravel SimpleTime | 0 |  |  |
|  | ZenoTravel Strips | 0 |  |  |
|  | ZenoTravel HC SimpleTime | 0 |  |  |
|  | ZenoTravel HC Strips | 0 |  |  |
|  | ZenoTravel HC Time | 0.017 |  |  |
| TLPlan | Depots Numeric | 0.0009 | Satellite HardNumeric | 1 |
|  | Depots SimpleTime | 0.01 | Satellite HC Complex | 1 |
|  | Depots Strips | 0.0009 | Satellite HC Numeric | 1 |
|  | Depots Time | 0.0003 |  |  |
|  | Depots HC Numeric | 0 |  |  |
|  | Depots HC Time | 0.0084 |  |  |
|  | DriverLog HardNumeric | 0.0051 |  |  |
|  | DriverLog Numeric | 0.0083 |  |  |
|  | DriverLog SimpleTime | 0.02 |  |  |
|  | DriverLog Strips | 0.0009 |  |  |
|  | Rovers HC SimpleTime | 0.0037 |  |  |
|  | Rovers HC Strips | 0.013 |  |  |
|  | Satellite SimpleTime | 0 |  |  |
|  | Satellite Strips | 0 |  |  |
|  | Satellite HC SimpleTime | 0.0001 |  |  |
|  | ZenoTravel SimpleTime | 0 |  |  |
|  | ZenoTravel Strips | 0 |  |  |
|  | ZenoTravel Time | 0.0014 |  |  |
|  | ZenoTravel HC SimpleTime | 0 |  |  |
|  | ZenoTravel HC Strips | 0 |  |  |
|  | ZenoTravel HC Time | 0 |  |  |

Figure 25: Easy/hard boundaries for hand-coded planners. Note: HC indicates the larger problems used only for the hand-coded planners.

STRIPS problems were significantly easy, by comparison with other STRIPS problems, for the majority of the participating planners, and not hard for any of them. On the other hand the Satellite NUMERIC problems were found to be challenging relative to other NUMERIC problems. Figure 23 shows that the hand-coded planners found ZenoTravel problems easy at all levels, by comparison with problems at similar levels, and this remains true for the large problem instances. Depots problems were also easy for the hand-coded planners.

When we consider the level-independent picture in the right-hand halves of Figures 22 and 23 we can observe that ZenoTravel emerges as significantly easy for the fully-automated planners, across all levels, by comparison with other problems irrespective of level. This pattern is broken by only one full-automated planner (LPG) finding these problems hard at the TIME problem level. The Satellite domain is similarly easy for the fully-automated planners, at all levels except NUMERIC. It can be noted that the number of planners finding the STRIPS problems easy in the level-independent comparisons is surprisingly high. The interpretation is that the problems in the population as a whole are much harder, so that the performance on STRIPS problems is pushed to the extremes of the performance on all problems. The hand-coded planners found the Depots and ZenoTravel problems to be uniformly easy at all levels.

Considering both the fully-automated and the hand-coded planners, the DriverLog, Rovers and Satellite domains present the most varied picture, suggesting that the problems in these domains presented the greatest challenges overall. All of the hand-coded planners found the SIMPLETIME Rovers problems significantly hard relative to other SIMPLETIME problems, but only one found these problems amongst the hardest that they had to solve overall. Interestingly, the perceived difficulty of the small Rovers problems does not persist into the large problems.

An interesting comparison can be made between the results of the analysis for STRIPS domains and the work of Hoffmann (2003b) analysing the topologies of STRIPS and ADL versions of the common planning benchmark domains. Hoffmann examines the behaviour of the $h^+$ function, measuring relaxed distances between states in the state spaces for these problems, in order to determine whether the function offers a reliable guide to navigate through the state space in search of plans. According to Hoffmann's analysis, the STRIPS versions of Depots, DriverLog and Rovers have local minima in the function and can have arbitrarily wide plateaus (sequences of states with equal values under $h^+$). These features can make problem instances in these domains hard for planners relying on $h^+$ (or approximations of it) to guide their search. This includes most of the fully-automated planners in the competition. However, interestingly, several of the fully-automated planners found one or more of these three domains to be easy at the STRIPS level (although in a few cases they were found to be hard). As Hoffmann points out, the potential hardness of a domain does not mean that all collections of problem instances from that domain are hard. Our observations seem to suggest that the competition collections posed instances that tended towards the easy end of the spectrum. This was unintentional and demonstrates that it can be difficult to obtain a good spread of challenges, particularly when generating problems automatically. Satellite and ZenoTravel domains have, in contrast, constant-bounded plateaus and therefore the $h^+$ function is a reliable guide in navigating the state space for these domains. Interestingly, in our analysis all fully-automated planners found these domains either easy or neither easy nor hard at the STRIPS level.

## 7. Scaling Issues

Section 6 addressed the issue of relative difficulty of problems *without* considering the question of whether planners agree about the difficulty of specific problems. The results of that section allow us to conclude that there is no overall consensus about which of the competition domains and levels were found hard, but it does not allow us to determine which planners agreed or disagreed on particular domains and levels. In order to look at the relative scaling behaviour of planners we need to identify the extent of such agreement. This is because to examine scaling behaviour it is necessary to have a scale that measures performance in a way that is meaningful to both planners in a comparison. The analysis described in this section therefore seeks to establish statistical evidence of such agreement.

In order to evaluate scaling behaviour we first explore whether the competing planners agree on what makes a problem, within a particular domain and level, hard. Although it might seem straightforward to ensure that a problem set consists of increasingly difficult problems (for example, by generating instances of increasing size) in fact it is not straightforward to achieve this. It appears that problem size and difficulty are not strongly correlated, whether size is taken as a measure of the number of objects, the number of relations or even the number of characters in a problem description. Although a coarse relationship can be observed — very large instances take more time to parse and to ground — small instances can sometimes present more difficult challenges than large instances. This indicates that factors other than size appear to be important in determining whether planners can solve individual instances.

In summary, the hypotheses explored in this section are:

> **Null Hypothesis:** The planners differ in their judgements about which individual problem instances are hard within a given domain/level combination.

> **Alternative Hypothesis:** The planners demonstrate significant agreement about the relative difficulties of the problem instances within any given domain/level combination.

In this section we are specifically concerned with a within-domain/level analysis and with whether planners agree on the relative difficulty of problem instances within a given domain/level combination.

### 7.1 Analytic Framework

As discussed in Section 6, we use the planners themselves as judges to determine how difficult individual problems were. Given that most of the competing planners proceeded by first grounding the problem instance and then by searching the problem space using some variation on the theme of a relaxed distance estimate, there seems little reason to believe that the planners would strongly diverge. If a particular instance, or family of instances, proved difficult for one planner it might be expected that this same collection would be challenging for all the competitors. To avoid being distracted by the impact of hand-coded control rules we separate the judgements of the fully-automated planners from those of the hand-coded planners. For each domain/level combination the hypothesis is that planners tend to agree about the relative difficulties of the problems presented within that domain and level.

| Fully Automated | Strips | Numeric | HardNumeric | SimpleTime | Time | Complex |
|---|---|---|---|---|---|---|
| Depots | $F_{21,110} = 5.3$ | $F_{21,44} = 5.48$ | | $F_{21,66} = 1.77$ | $F_{20,63} = 2.14$ | |
| DriverLog | $F_{19,100} = 17.1$ | $F_{19,40} = 17.4$ | $F_{19,40} = 4.05$ | $F_{19,60} = 4.44$ | $F_{19,60} = 4.63$ | |
| ZenoTravel | $F_{19,100} = 21.7$ | $F_{19,40} = 14$ | | $F_{19,60} = 9.4$ | $F_{17,36} = 12.1$ | |
| Rovers | $F_{19,80} = 4.54$ | $F_{18,38} = 9.47$ | | $F_{19,60} = 4.25$ | $F_{19,60} = 6.92$ | |
| Satellite | $F_{19,100} = 7.36$ | $\mathbf{F_{15,48} = 1.74}$ | $F_{19,20} = 11.8$ | $F_{19,60} = 3.6$ | $F_{19,60} = 4.19$ | $F_{19,60} = 3.78$ |
| FreeCell | $F_{19,100} = 6.21$ | | | | | |
| Settlers | | $\mathbf{F_{5,6} = 1.6}$ | | | | |

| Hand-Coded (Small) | Strips | Numeric | HardNumeric | SimpleTime | Time | Complex |
|---|---|---|---|---|---|---|
| Depots | $F_{21,44} = 2.49$ | $F_{21,22} = 2.19$ | | $F_{21,44} = 4.54$ | $F_{21,44} = 6.21$ | |
| DriverLog | $F_{19,40} = 2.58$ | $F_{19,20} = 3.73$ | $F_{19,20} = 6.45$ | $F_{19,40} = 5.34$ | $F_{19,40} = 6.52$ | |
| ZenoTravel | $F_{19,40} = 2.93$ | $F_{19,20} = 8.3$ | | $F_{19,40} = 5.54$ | $F_{19,40} = 4.22$ | |
| Rovers | $F_{19,40} = 4.5$ | $F_{19,20} = 36.5$ | | | $F_{19,40} = 18$ | |
| Satellite | $F_{19,40} = 7.25$ | $F_{19,20} = 38.7$ | $F_{19,20} = 9.4$ | $F_{19,40} = 5$ | $F_{19,40} = 20.6$ | $F_{19,20} = 51.3$ |

| Hand-Coded (Large) | Strips | Numeric | HardNumeric | SimpleTime | Time | Complex |
|---|---|---|---|---|---|---|
| Depots | $F_{21,44} = 11.4$ | $F_{21,22} = 3.76$ | | $F_{21,44} = 13.8$ | $F_{21,44} = 10.8$ | |
| DriverLog | $F_{19,40} = 61.4$ | $F_{19,20} = 57.3$ | $F_{19,20} = 66.4$ | $F_{19,40} = 91.6$ | $F_{19,40} = 80.5$ | |
| ZenoTravel | $F_{19,40} = 3.14$ | $\mathbf{F_{19,20} = 1.47}$ | | $F_{19,40} = 3.54$ | $F_{19,40} = 3.37$ | |
| Rovers | $F_{19,40} = 17.2$ | $F_{19,20} = 29.9$ | | $F_{19,40} = 33$ | $F_{19,40} = 49.4$ | |
| Satellite | $F_{15,32} = 20.8$ | $F_{15,16} = 33.5$ | | $F_{15,32} = 43.7$ | $F_{15,32} = 88.5$ | $F_{15,16} = 152$ |

Figure 26: F-values for the multiple judgments rank correlation tests.

To explore the extent to which agreement exists we perform rank correlation tests for agreement in multiple judgements (Kanji, 1999) (we refer to this test as an MRC). In our experiment the judges are the planners and the subjects are the problem instances. We perform a distinct MRC for each domain/level combination, showing in each case how the planners ranked the instances in that domain and level. We therefore perform 25 MRCs for the fully-automated planners (there were 25 distinct domain/level pairs in which the fully-automated planners competed), 23 for the hand-coded planners on the small problems (the hand-coded planners did not compete in the Freecell STRIPS or Settlers NUMERIC domains) and 22 for the hand-coded planners on the large problems (amongst which there were no Satellite HARDNUMERIC instances). The results of these tests are shown in Figure 26. In each test the $n$ planners rank the $k$ problem instances in order of time taken to solve. Unsolved problems create no difficulties as they are pushed to the top end of the ranking. The MRC determines whether the independent rankings made by the $n$ planners agree. The test statistic follows the F-distribution with $(k-1, k(n-1))$ degrees of freedom determining whether the critical value is exceeded.

## 7.2 Results of Analysis

The cells in Figure 26 report the F values obtained (and the degrees of freedom used). In almost all cases the critical value was exceeded and the null hypothesis of non-agreement could be rejected for at least the 0.05 level. In just a few cases (those reported in bold font) the critical value was not exceeded and no statistical evidence was therefore found of agreement between the planners about the difficulty of instances in the corresponding domain and level. It is interesting to note that the problematic cases are all within the NUMERIC level, for both fully-automated and hand-coded planners. Furthermore, the case that comes closest to the critical boundary (the small Depots NUMERIC problems, in the hand coded table, with an F-value of 2.19) is also within the NUMERIC level.

## 7.3 Interpretation

The results support rejection of the null hypothesis in almost all cases. We can therefore adopt the alternative hypothesis, observing that there are many cases in which planners do agree on the relative difficulties of problem instances within a given domain/level combination.

During the competition we observed that TALPLANNER is at a disadvantage with respect to the other hand-coded planners, in terms of comparative speed, when running on small problems. This is probably because of the java virtual machine start-up time which becomes significant relative to actual solution time on small instances. We see the effects of this start-up time in the tables. Note that, in those domain/level combinations in which TALplanner competed (STRIPS, SIMPLETIME and TIME) we see a low level of agreement amongst the hand-coded planners on the small problems (except in the case of the Rover domain). This is not because TALPLANNER disagrees with the other planners about the ranking of the actual problems, but because the problems are small enough that the variability in setup time throws noise into the ranking and obscures the true picture of relative problem difficulty. With the set of large problems we see that this anomaly is removed — the problems are sufficiently challenging that the java startup time becomes insignificant — and a high level of agreement over ranking is obtained. Interestingly, the hand-coded planners show a consistently high level of agreement about the ranking of Rovers problems. The fact that this does not emerge in the fully-automated set may be due to the larger number of judges in the fully-automated category.

## 8. Relative Scaling Behaviours

The MRCs described in Section 7 demonstrate that the planners do agree, as expected, about the relative difficulty of problem instances *within* most domain/level combinations. In these cases it is possible to go on to explore the domain and level specific scaling behaviour of the planners, and we go on to investigate that in this section. We cannot explore the scaling behaviour of the planners across domains because, as we discussed in Section 6, there does not seem to be much across-the-board agreement concerning the relative hardness of the domains so we would be unlikely to see agreement in multiple judgments across the domain boundaries.

The ideal basis on which to explore scaling behaviour would be to have a collection of problems with a canonical scaling of difficulty and then to compare the performance of planners as they scaled on progressively harder problems within this collection. Unfortunately, many factors contribute to making problems hard and these do not affect planners uniformly. As a result, there is no canonical measurement of problem difficulty in many domains. Instead, we must determine the relative difficulty of problems by using the planners themselves as judges. This means that we can only consider the relative scaling behaviours of planners when the planners agree on the underlying ordering of the difficulty of problems. Thus, we begin by identifying appropriate sets of problems — those on which a given pair of planners agree about the relative hardness of problems according to our analysis in Section 7 — and then proceed to compare the way that each of the planners in the pair scales as the problems increase in difficulty. The first stage of the analysis considers only the order that the two planners place on the problems within a set, while the second stage examines how the performance varies between the two planners as they progress from problem to problem.

The hypotheses explored in this section are:

>**Null Hypothesis:** Where planners agree about the difficulty of problems for a given domain/level combination, they exhibit the same scaling behaviour.

>**Alternative Hypothesis:** Where planners agree about the difficulty of problems for a given domain/level combination, they demonstrate different scaling behaviours, where the better scaling performance can be identified from the data set.

This section is concerned with the question of scaling behaviour within problem sets from specific domain/level combinations in which there is already determined to be agreement, as identified in Section 7.

## 8.1 Analytic Framework

In order to test the different scaling properties of the planners we make pairwise comparisons of performance using only those domains where both planners agreed about the difficulty of the problems. That is, we use a domain in a comparison if both planners found it hard, both found it easy or neither found it hard or easy.

To rank the problems in order of difficulty we use the results obtained from the bootstrapping experiment described in Section 6. Our rankings are level dependent, so we looked at scaling within the four problem levels and recorded the results separately. We do not attempt to combine these results into a single overall conclusion about scaling — we recognize that different planners scale better at some problem levels than at others, and that no single planner can therefore emerge as scaling best overall.

We only compare two planners if they agreed about difficulty in at least two domains. This gives us, in each case where a comparison is made, a data set of more than 30 points. Where the planners did not agree in at least two domains we conclude that there is insufficient agreement between them about what constitutes problem difficulty for it to be possible to measure their relative scaling behaviours in a meaningful way.

To perform a comparison between two planners we rank the problems in the data set in order of agreed difficulty and then rank the differences between the performances of the planners on these problems. We then explore whether the ranking of the differences is correlated with the ranking of the problems according to their difficulty. We use ranks because we cannot make assumptions about the shapes of the underlying problem distributions or the functions that truly describe the performances of the planners being compared, so our results are robust with respect to these factors.

| | FF | LPG | | MIPS | | | | Sapa | VHPOP | |
|---|---|---|---|---|---|---|---|---|---|---|
| | | STRIPS | NUMERIC | STRIPS | NUMERIC | SIMPLE TIME | TIME | TIME | STRIPS | SIMPLE TIME |
| FF | | 0.36 | ⊙ | 0.87 | 0.93 | ——⊗—— | | ⊗ | 0.93 | ⊗ |
| LPG | | | | ⊙ | 0.52 | 0.51 | 0.61 | 0.58 | 0.44 | 0.48 |
| MIPS | | | | | | | | ⊙ | ——⊙—— | |
| Sapa | ⊗ | | | ⊗ | ——————⊙—————— | | | | ——⊙—— | |
| VHPOP | | | | ⊙ | ⊗ | ⊙ | ⊗ | ⊗ | | |

Figure 27: Table showing correlation values, for fully-automated planners, between problem difficulty and difference in time performance, indicating scaling behaviour. ⊗ means that one of the pairs of planners did not produce data so no comparison may be drawn. ⊙ means that there was insufficient agreement between the planners on the difficulty of domains or the ranking of problems in order to carry out a comparison.

Given two planners, $p_1$ and $p_2$, a positive correlation between the rankings of the differences in values between $p_1$ and $p_2$ and the problem difficulty ranking means that the difference in performance between $p_1$ and $p_2$ (that is, performance of $p_1$ minus performance of $p_2$) is increasing as the problems become more difficult. If the curve of $p_1$ is increasing faster $p_2$ scales better than $p_1$. A negative correlation means that $p_1$ scales better than $p_2$. A zero (or near-zero) correlation means that the scaling behaviour of the two planners is insignificantly different. We use Spearman's rank correlation test (see Appendix C) to identify the critical value required for confidence at the 0.05 level.

We restrict our attention to the planners that solved the most problems overall in the two categories. Thus, the fully-automated planners we compared were FF, LPG, MIPS, VHPOP and SAPA. We consider all pairs of the hand coded planners. We do not perform any cross-category tests as it is evident from the raw data that the hand coded planners exhibit better scaling behaviour than any of the fully-automated planners.

## 8.2 Results of Analysis

The table in Figure 27 shows the significant scaling differences that we found between pairs of fully-automated planners at each of the levels. Figure 28 shows the relative scaling of the hand coded planners. In both sets of tests, two planners could only be compared at the levels at which both competed, and on domains in which they both agreed were either easy, hard, or neither easy nor hard. We report the results so that the planner indexed by row is the one showing the superior scaling behaviour. Where planners did not compete in the same tracks we indicate this with the symbol ⊗ denoting *incomparable*. Where no significant difference in scaling was found we indicate this with a zero correlation. Where no agreement was found to support a comparison we use the symbol ⊙ denoting *disagreement*. To avoid duplication of data, we place entries as positive correlations only in the cell corresponding to the row for the planner favoured by the comparison and omit the corresponding negative correlation in the cell for which row and column planners are reversed.

|  | TLPlan | | SHOP2 | | | | TALPlanner | | |
|---|---|---|---|---|---|---|---|---|---|
|  | STRIPS | TIME | STRIPS | NUMERIC | SIMPLE TIME | TIME | STRIPS | SIMPLE TIME | TIME |
| TLPlan |  |  | 0.77 | 0.93 | 0.85 | 0.83 | 0 | 0.25 | 0 |
| SHOP2 |  |  |  |  |  |  |  |  |  |
| TALPlanner | 0 | 0 | 0.86 | ⊗ | 0.46 | 0.76 |  |  |  |

Figure 28: Table showing correlation values, for hand-coded planners, between problem difficulty and difference in time performance, indicating scaling behaviour. ⊗ means that one of the pairs of planners did not produce data so no comparison may be drawn.

## 8.3 Interpretation

In almost all cases in which a comparison could be performed, a significant difference in scaling behaviour was found, supporting rejection of the null hypothesis.

Because we used only those domains where there was agreement about the relative difficulty of the problems it is not necessary to restrict our conclusions to be domain-dependent. However, we had only a restricted collection of data points at our disposal so we must be careful how we generalise this picture. On the basis of our analyses we believe we can make some tentative judgements about how planners are scaling in pairwise comparisons within the four competition levels.

For the fully-automated planners it can be observed informally that there is a high degree of consistency between the scaling behaviours of particular planners across the problem levels in which they competed. Although we cannot draw overall conclusions from the data set with a high level of confidence we can observe that FF exhibits the best scaling behaviour in the levels at which it competed and LPG exhibits the best scaling behaviour at the temporal levels. It should be remembered that we did not perform single-domain comparisons, although these might be interesting from the point of view of exploring domain-specific scaling behaviour and might produce some interesting results. We felt that these results would be interesting curiosities rather than anything that could support general conclusions.

The hand coded planners also show a high degree of cross-level consistency. It can be observed informally that TLPLAN scales much better than SHOP2 across all levels, whereas it scales only marginally better than TALPLANNER in the STRIPS domains and not significantly in any other level. TALPLANNER scales better than SHOP2 at all levels in which they both competed. It can be seen that SHOP2 is not scaling well relative to its competitors, although it should be remembered that the quality of plans produced by SHOP2 is superior in some domains.

Formally the tables allow us to draw specific conclusions about the relative scaling behaviours of specific pairs of planners, within specific problem levels, at the 0.05 level.

## 9. Conclusions

The 3rd International Planning Competition focussed on the issue of temporal planning and numeric resource handling. The competition was structured around two categories:

fully-automated and hand-coded planning systems, and four main problem levels: STRIPS, NUMERIC, SIMPLETIME and TIME. There were eight domains, one of which was intended for the hand coded planners only (the UM-TRANSLOG domain), and two solely for the fully-automated planners (the FreeCell and Settlers domains). Fourteen competitors took part, eleven in the fully-automated track and three in the hand coded track. The domain description language used was PDDL2.1, an extension of the PDDL standard designed for modelling temporal and resource-intensive domains. PDDL2.1 is described in another paper in this issue (Fox & Long, 2003).

We collected a data set of some five and a half thousand data points distributed over the domains and levels. An initial plotting of these points, in terms of the relative time and quality performances of the planners in the different domains, revealed a number of interesting patterns. These suggest some characteristics of the relative performances of the competitors *within* the competition domains. These patterns were presented and discussed at the AIPS conference with which the final competition events were co-located. However, other patterns, such as those indicating relative performances across domains and those showing the perceived difficulty of the competition domain/level combinations, were invisible in the data presented in this way. This paper presents the results of some detailed statistical analyses of the competition data, aimed at identifying some of these deeper patterns.

This paper explores three experimental themes. The first theme is aimed at answering the question: *which planner should I buy?* This question is concerned with which planner emerges as the strongest overall performer, rather than which produced the best results in a particular domain/level combination, and it can be asked from the point of view of both speed and quality criteria. To answer it we performed comparisons, based on the Wilcoxon rank-sum matched-pairs test, enabling the construction of partial orders on the competing planners in terms of their time and quality performances in the four levels of the competition. From these partial orders it can be concluded that, if a potential user is interested in good behaviour across a broad range of temporal and numeric problems then LPG, amongst the fully-automated planners, and TLPLAN, amongst the hand-coded planners, are the best choices. Of course, if more specialised coverage is required and speed of solution is paramount then other choices might be made.

The second theme considers the dependence of planner performance on domain structure. We were interested in exploring the extent to which the competing planners agree about which domain/level combinations were hard and which were easy. The analysis we performed in addressing the first of these issues is a statistical complement to the theoretical analysis of domain topologies being carried out by Hoffmann (2003b). We considered the competition domains at all four levels used in the competition, whilst Hoffmann considers only the STRIPS subset of the competition domains (he also considers ADL domains, but we did not use any of these in the competition). It is interesting to note that our findings were broadly consistent with his conclusions.

The third theme considered the scaling behaviour of the competing planners. We considered two related issues: the extent to which the competing planners agreed on the relative difficulty of the problem instances within domain/level combinations and the extent to which the planners scaled similarly in domain/level combinations where there was agreement. Our intentions in pursuing the first issue were to provide an objective scale that would support our efforts to investigate the relative scaling behaviours of the planners. Because we found relatively little agreement over the perceived difficulty of problems within domain/level combinations we were able to perform only a restricted comparison of relative scaling behaviour. However, we consider the results we obtained to make an interesting contribution to a deeper comparison of planner performances than is available from the raw domain-specific data.

There are many other questions that it would be interesting to be able to answer. Amongst these are questions about the extent to which hand-coding of domain knowledge really benefits a planner and the amount (and value) of effort involved in encoding such knowledge. This is a pressing question for the community and one that the competition series might be well-suited to try to answer. However, in order to pursue these in future competitions it will be necessary to carefully design controlled experiments aimed at exploring the precise hypotheses. We have been restricted in this paper to the post hoc analysis of the competition data, and this has clearly restricted the kinds of questions we have been able to ask and answer. However, we hope that both the results and the methodologies presented here will be of interest to the planning community and that they will help to encourage further scientific evaluation of performance in the field.

## Acknowledgements

We would like to thank all the competitors in the 3rd International Planning Competition for contributing their time and enthusiasm to the event we report and for providing the data that has made this paper possible. We would also like to thank Adele Howe who has contributed invaluable advice, comments and huge support and made it possible for this paper to go far further than we could have hoped. We would like to thank David Smith for undertaking the unenviable task of coordinating and editing this special issue of the Journal of Artificial Intelligence Research and doing so with immense patience, good humour and generous support. Finally, we would like to thank Martha Pollack who first proposed the idea of publishing the assembled work in a special issue of the Journal of Artifical Intelligence Research. Her whole-hearted commitment to the project has been vital to its successful conclusion.

## Appendix A. Problem Domains

### A.1 The First International Planning Competition

The first competition used the following domains:

- **Logistics** A transportation problem involving aircraft and trucks, with trucks constrained to movement within cities and aircraft constrained to movement between (inter-city) airports. This domain allows considerable parallelism.

- **Mystery** A transportation domain with vehicles having limited capacity and consuming limited stocks of fuel.

- **MPrime** A variant of the Mystery domain in which it is also possible to pipe fuel between locations in order to allow vehicles to have different movement options.

- **Grid** A problem in which a single robot moves between locations on a grid shaped map. Locations may be locked and there are keys that must be collected to gain access to these locations. The objectives of the problem instances involve transporting the keys to particular locations.

- **Gripper** A simple domain, originally designed to demonstrate the limitations of Graphplan, in which a collection of identical balls must be transported by a robot with two grippers from one room to an adjacent room.

- **Movie** A simple domain intended to explore use of conditional effects. A collection of snacks must be assembled prior to rewinding a video and then watching the movie.

- **Assembly** A complex ADL domain with a challenging use of quantified and conditional effects.

### A.2 The Second International Planning Competition

The second competition introduced several new domains:

- **Blocks** The classic blocks-world problem, encoded without an explicit reference to a gripper. This domain has significant goal interaction.

- **Job-Schedule** A problem involving the machining of parts. This problem exercises the ADL features involving conditional and quantified effects, although it is less complex than the Assembly domain.

- **Freecell** This is the classic solitaire card game that is widely available as a computer game. The encoding as a STRIPS domain represents a larger problem than most previous benchmarks and includes the awkward addition of an encoded set of integers.

- **Miconics Elevator** This domain was inspired by the problem of planning an efficient call sequence for an elevator car travelling between floors of a large building. There were several variants, with the most complex including numeric preconditions as well as purely logical constraints. An ADL version offered complex preconditions involving several different connectives and a STRIPS version offered a relatively simple transportation problem.

In addition, the Logistics domain was reused to provide some calibration of performance in comparison with the first competition.

## A.3 The Third International Planning Competition

### A.3.1 The Depots Domain

The domain consists of actions to load and unload trucks, using hoists that are available at fixed locations. The loads are all crates that can be stacked and unstacked onto a fixed set of pallets at the locations. The trucks do not hold crates in a particular order, so they can act like a table in the Blocks domain, allowing crates to be reordered.

This domain was devised with the foremost intention of testing strips planners. The second competition had demonstrated that the Logistics domain was no longer a serious challenge, and that, for planners using hand-coded controls, the Blocks domain was also solved. For fully-automated planners the Blocks domain still represents a challenge, although the second competition showed that some planners can solve quite large problems (up to twenty blocks) with reasonably efficient plans within a few minutes. However, performance can vary widely and there are problems in this range that can prove unsolvable for these planners. We wanted to see whether the performance that had been achieved in these domains could be successfully brought together in one domain. We were interested to see for fully-automated planners, where the interaction between the problems creates an additional family of choice points in addition to those that appear in the transportation and the block-tower-construction sub-problems. We were also interested to see this for hand-coded planners where the rules for each of the sub-problems are obviously well-understood, but it is not obvious whether the rules can be combined into a single collection without any problems of interaction.

The metric version of the domain adds weight attributes to the crates and weight capacities to the trucks. In addition, trucks consume fuel in their travels and the plans must minimise fuel use. Fuel use is constant and not dependent on the locations. Fuel is also consumed in lifting crates, so there is a tradeoff to be considered when crates must be restacked at a location. Either a truck can be brought in to act as a "table" or more complex lifting and stacking can be performed using the locally available pallets as transfer space.

The temporal versions allow for concurrent activities of the trucks and the hoists (at all locations). The full temporal variant makes the time for driving dependent on the truck and the distance between the locations, and makes the time to load or unload a crate dependent on the weight of the crate and the power of the hoist. The objective in both is to minimise make-span (the overall duration of the plan).

### A.3.2 The DriverLog Domain

This domain has drivers that can walk between locations and trucks that can drive between locations. Walking requires traversal of different paths from those used for driving, and there is always one intermediate location on a footpath between two road junctions. The trucks can be loaded or unloaded with packages (with or without a driver present) and the objective is to transport packages between locations, ending up with a subset of the packages, the trucks and the drivers at specified destinations.

This domain was produced to explore the power of STRIPS solutions to transportation problems when the transportation involves a sub-problem of acquiring a driver. The problem is one that offers significant opportunity for concurrency in the use of drivers and vehicles, so we were also interested to see how the temporal variants were handled.

The metric variant of the domain adds costs for walking and driving and problem instances required that the planner optimise some linear combination of the total walking cost and the total driving cost.

The full temporal variant makes time spent driving or walking between locations dependent on the path being traversed, but other durations are only dependent on the actions (as with the simple temporal version). In both of these variants the plan quality depends on the make-span.

An additional variant, the hard numeric variant, complicates the cost of driving by making it dependent on the load being carried: each additional package added to a truck increases the fuel consumption rate of the truck by its current value, making the consumption increase as a quadratic function of the load.

### A.3.3 The Zeno-Travel Domain

This domain has actions to embark and disembark passengers onto aircraft that can fly at two alternative speeds between locations. The STRIPS variant is rather uninteresting because the two speeds do not offer meaningful alternatives. In the metric variant the planes consume fuel at different rates according to the speed of travel (two alternatives) and distances between locations vary. Problem instances require plans to minimise some linear combination of time and fuel use.

The temporal versions are closer to the original ZENO problem. They involve durations for the different means of travel and different levels of fuel consumption. In contrast to the original ZENO problem the fuel consumption is not described by a continuous function, but by discrete step functions applied at the end points of durative actions. The fact that the fuel in an aircraft cannot be affected by multiple different concurrent actions and its value is not relevant to satisfying the precondition of any actions that could begin during the continuous consumption or replenishment of fuel means a discrete model of fuel use is sufficient, demanding less expressive power of the planners that use the model.

### A.3.4 The Satellite Domain

The satellite domain was developed following discussions with David E. Smith and Jeremy Franks at NASA Ames Research Center. It is intended to be a first model of the satellite observation scheduling problem. The full problem involves using one or more satellites to make observations, collecting data and downlinking the data to a ground station. The satellites are equipped with different (possibly overlapping) collections of instruments, each with different characteristics in terms of appropriate calibration targets, data productions, energy consumption and requirements for warming up and cooling down. The satellites can be pointed at different targets by slewing them between different attitudes. There can be constraints on which targets are accessible to different satellites due to occlusion and slewing capabilities. Instruments generate data that must be stored on the satellite and subsequently downlinked when a window of communication opportunity opens with a ground station. Communication windows are fixed. Data takes time to downlink and it could be impossible to downlink an entire satellite store in a given time frame, so downlinks must be scheduled around the storage capacity, the production of data by observations and the opportunities to downlink data as they arise. In the real problem there are additional difficulties such as the management of energy and the use of solar power and the maintenance of operational temperatures during periods in shadow. In order to make the problem accessible to the planners in the competition (given the time scales for encoding the domain, writing a problem generator and testing competing planners) several important features of the real problem are simplified. Perhaps most important of these is that in the real problem targets are only visible during particular time-windows, although the elimination of the problem of downlinking data is also a significant simplification. Representing the windows of opportunity is possible in PDDL2.1, but not entirely straight-forward and this remains an area in which there is need for development. Management of power and temperature were also simplified away.

The STRIPS version of the problem involves deciding on the most efficient covering of the observations given the satellite capabilities. This is an interesting combinatorial problem if the satellites are assumed to be free to operate concurrently (in a Graphplan-style parallel activity), but otherwise the problem offers few interesting choice points. The STRIPS version was based on an earlier Satellite domain contributed by Patrik Haslum.

The metric version of the problem introduces data capacities into the satellites and fuel use in slewing between targets. The plans are expected to minimize fuel use in obtaining the data. This problem combines a constrained bin-packing problem (getting the data into the limited stores of the satellites, subject to the constraints that only certain satellites are equipped to obtain certain data) with a kind of route planning problem (finding fuel-efficient paths between targets while also considering the combined costs of the fuel consumption by all of the satellites).

The temporal versions introduce duration and make concurrency important. The full temporal problem includes different slew times between different pairs of targets. These problems both involve minimising make-span times for the data acquisition. A complex version of the domain combines the temporal and metric features so that planners are required to manage the problem of storing different sized data blocks in limited capacity satellite data stores.

A further variant of the Satellite domain, called the HARDNUMERIC version, represents an important departure from traditional planning problems: the logical goals describing the intended final state are trivial (either empty or a few simple final positions for the satellites), but the metric by which the plans are declared to be evaluated is the quantity of data collected. The problem is interesting because a null (or nearly null) plan will solve all the instances, but the quality of these plans will be zero. To produce a *good* plan it is necessary to ensure that the satellites are used to collect data and this, in turn, requires that the planner constructs reasonable data collection goals. This is a hard problem for most current planners — particularly the fully-automated planning systems. Nevertheless, it is a realistic demand for many problems that a planner might be required to face: it is not uncommon for the specific final state to be less important than the effects of the actions carried out in reaching it.

### A.3.5 THE ROVERS DOMAIN

The Rovers domain was constructed as a simplified representation of the problem that confronts the NASA Mars Exploration Rover missions launched in 2003, the Mars Science Laboratory mission planned for 2009 and other similar missions that are expected as part of the ESA AURORA project. The STRIPS version of the problem involves planning for several rovers, equipped with different, but possibly overlapping, sets of equipment to traverse a planet surface. The rovers must travel between waypoints gathering data and transmitting it back to a lander. The traversal is complicated by the fact that certain rovers are restricted to travelling over certain terrain types and this makes particular routes impassable to some of the rovers. Data transmission is also constrained by the visibility of the lander from the waypoints.

The metric version of the domain introduces an energy cost associated with actions and an action allowing rovers to recharge, provided they are in the sun. The problems sought solutions that minimised numbers of recharges, so the use of energy was required to be as efficient as possible.

In the metric temporal variant the domain involves both energy and time management, although the instances require planners to optimise total duration. This demand implies the need for efficient energy use, since recharging is an action that consumes a considerable amount of time and, in our model, requires the recharging rover to remain in one place during the period of recharging. The opportunity for careful division of labour between the rovers makes both temporal variants complex and interesting problems.

### A.3.6 THE SETTLERS DOMAIN

This domain exists only as a metric problem. The problem is inspired by the multitude of computer games that involve managing resources, accumulating raw resources and slowly combining them into more and more advanced processing plants and other structures to achieve more sophisticated objectives. The problem was proposed by Patrik Haslum. An interesting difficulty that the problem presents is that it is a problem that involves constructing new objects out of resources. This is not easily represented in PDDL2.1. In fact, the only way to capture the domain in PDDL2.1 is to name the objects that *could* be constructed and to give them an attribute indicating whether or not they have actually been constructed. This leads to a very cumbersome encoding and, moreover, represents a particular problem for planners that ground actions prior to planning, since a multitude of potential objects must be considered in the process, leading to a huge collection of actions. Many of these actions are uninterestingly different, since the specific choice of names for objects that are created in solving an instance is clearly irrelevant, but each alternative is constructed and considered in the planning process.

### A.3.7 THE UM-TRANSLOG-2 DOMAIN

The UM-Translog-2 (Wu & Nau, 2002) domain was used only for the hand-coded planners and only an incomplete set of results was collected due to time constraints at the conclusion of the testing period. This domain is a significant challenge because of its size and should be seen as a useful benchmark problem for fully-automated planners because of the challenges in both grounding and searching a domain with so many action schemas.

## Appendix B. The Competitors

### B.1 The Fully Automated Planners

There were eleven competitors in this category, representing at least four distinct planning paradigms (forward search, model-checking, local search and partial order planning). Fully-automated planners accept the PDDL2.1 specifications of domain, initial state and goal and compute solutions solely on the basis of these specifications. No additional control knowledge or guidance is supplied. Fully-automated planners therefore depend on sophisticated search control heuristics and the efficient storage of alternative search branches. A popular current approach to search control is to make use of variants of the relaxed plan idea originally proposed by McDermott (1996) and subsequently exploited by Bonet and Geffner (1997) and Hoffmann (2000).

IxTeT (Laborie & Ghallab, 1995) entered as a fully-automated planner, but in retrospect it might have been better classified as a hand-coded planner. The ability to hand-code domain representations might have alleviated some of the problems that arose in making the competition domains accessible to IxTeT. IxTeT does not currently accept domains or problems in PDDL format, so it was necessary to translate the competition domains into its own representation language. No automatic translator between PDDL and the IxTeT input language yet exists and it is not clear that such a translation can be automated. Furthermore, the plan representation of IxTeT is more general than that insisted upon for use in the competition but this was not an advantage given the need to automate the plan validation process. In fact, several plans produced by IxTeT could not be validated. The combination of these difficulties made it impossible to offer any real insights into the performance of IxTeT on the competition domains. Nevertheless, we were pleased that an attempt was made to enter IxTeT: it is important that the competition should not cause a fracture between the members of the research community who are interested in entering and those who have long-established alternative planning technology that cannot be easily reengineered to meet the assumptions underlying the competition.

#### B.1.1 FF

FF has been an extremely successful and influential planner since 2000 (Hoffmann & Nebel, 2000). It is based on forward state space search using relaxed plans to give heuristic guidance in its choice between possible steps through the space. Hoffmann extended the original FF system (Hoffmann, 2003a) to include a treatment of metric domains by relaxing the metric dimensions of the problem as well as the logical dimensions.

#### B.1.2 IxTeT

IxTeT (Laborie & Ghallab, 1995) is well known for having been one of the first planners to reason about time and resource intensive domains. The version that participated in parts of the competition is a reimplementation of the original system described by Ghallab and Laruelle (Ghallab & Laruelle, 1994). As mentioned above, IxTeT experienced a number of difficulties in the competition making it difficult to evaluate its performance. However, seen in the broader context of planning research and application IxTeT has made many important contributions to the development of temporal and resource-intensive planning approaches and, with its powerful plan representation language, is suited to certain applications for which the simplified plan representation used in the competition is inadequate.

### B.1.3 LPG

LPG (Gerevini et al., 2003) is based on a local-search strategy applied to plan graphs (Blum & Furst, 1995). The approach has been generalised to accommodate both metric and temporal structure, making it a powerful and flexible planner. The use of local search allows the planner to be configured to trade-off time and plan quality. Indeed, the planner exhibits any time behaviour in the sense that plans can be reported as they are found and, if the search is allowed to run longer, better quality plans might be discovered.

### B.1.4 MIPS

MIPS (Edelkamp, 2003) uses a variety of techniques, but at the core is a model-checker based on ordered binary decision diagrams (OBDDs), which is used to generate reachable states. The planner uses a powerful technique to compress state representations in order to make the OBDDs more compact. Exhaustive search of the state space is impractical in large problems and MIPS uses a heuristic evaluation function based on relaxed plans in order to restrict the space of explored states. MIPS tackles concurrency in temporal planning by lifting partial orders from the totally ordered plans that are produced by its forward search. MIPS has also been extended to manage metric quantities, also using a relaxation heuristic to predict the behaviours of metric quantities.

### B.1.5 SAPA

SAPA (Do & Kambhampati, 2003) is a forward search planner using a relaxed temporal plan heuristic (based on the use of a relaxed TGP-style (Smith & Weld, 1999) plan graph) to guide its search. The heuristic is supplemented with a heuristic estimate of resource usage allowing the planner to handle metric quantities. Temporal structure is managed using delayed effects, so that, when a durative action is executed, its end effects are queued in an event queue, pending application when time is advanced to the point at which they are triggered. The focus of SAPA development has been in the management of metric and temporal structure. SAPA did not attempt to compete in STRIPS or SIMPLETIME problems, but performed well in the more complex problems.

### B.1.6 SEMSYN

SEMSYN (Parker, 1999) is a Graphplan-based planner, with extensions to handle metric and ADL features. In general, Graphplan-based approaches have, with the exception of LPG, proven unequal to the challenge of scaling to meet the latest sets of benchmark problems. This suggests that the search strategy of Graphplan must be abandoned if large problems are to be solved, but that the underlying plan graph structure need not be a source of scaling problems (in fact FF, VHPOP, SAPA and LPG all use plan graph structures in the planning process).

### B.1.7 SIMPLANNER

SIMPLANNER (Onaindía, Sapena, Sebastia, & Marzal, 2001) is a forward search planner using a relaxed plan heuristic. The heuristic evaluation uses separate relaxed plans for each of the top level goals, combining them to identify a useful first action to apply from the current state. This variant on the idea of relaxed plans appears to represent a reinforcement of the notion of helpful actions developed in FF, where actions that are selected in the first layer of a relaxed Graphplan-style plan are favoured as appropriate candidates for the next step in a plan.

### B.1.8 STELLA

STELLA (Sebastia, Onaindía, & Marzal, 2001) uses a forward heuristic search architecture, but with the modification that plans are built using landmarks (Porteous, Sebastia, & Hoffmann, 2001). The idea is to identify key states in the path of a plan before planning begins and then to use these as "stepping stones" to progress from the initial state to the goal state.

### B.1.9 TP4

TP4 (Haslum & Geffner, 2001) is a development of the HSP (Bonet et al., 1997) planning approach, which was one of the first of the current generation of rather successful heuristic state-space search planners based on relaxed plan heuristics. TP4 extends the use of the heuristic to manage temporal plan structure. The planner is intended to find optimal plans (although, for minor technical reasons, some of the plans it produces are slightly sub-optimal), by using an admissible heuristic, and this requires a far greater search effort than in planners constructing plans that are merely heuristically good.

### B.1.10 TPSYS

TPSys (Garrido, Onaindía, & Barber, 2001; Garrido, Fox, & Long, 2002) is a temporal planner based on Graphplan. There are technical differences between TPSys and TGP (Smith & Weld, 1999), but the central use of a temporal plan graph is similar, using the graph to represent the passage of time, with actions having associated durations. As with other Graphplan-based approaches, the search machinery appears to scale badly.

### B.1.11 VHPOP

Partial order planners have suffered a period of being unfashionable, supplanted by Graphplan and, more recently, relaxed-plan-based heuristic state-space search planners. VH-POP (Younes & Simmons, 2003) represents an interesting indication that partial order planning is far from defunct. In particular, the partial order framework offers a powerful way to handle temporal constraints. In VHPOP a simple temporal network is used to manage the temporal constraints between the end points of durative actions and this allows the planner to successfully treat concurrency and other features of temporal plan structure. Within the framework of a partial-order planner, VHPOP makes use of plan graph distance estimates to guide its search.

## B.2 The Hand-coded Planners

There were three entrants in the category of planners requiring hand-coded control knowledge. As in the 2000 competition the teams competing with these planners were allowed time to reformulate domain descriptions to include domain-specific control knowledge. As the results show, control knowledge can dramatically improve planner performance. However, it is difficult to assess the cost-effectiveness of hand-coding control knowledge. Fahiem Bacchus (2001) observed the need to quantify the time and effort required to identify and encode useful control knowledge in order to be better able to judge the trade-off between the fully-automated and hand-coded approaches. However, it is very difficult to measure the effort involved. In principle it should be possible to bound the time allowed for domain reformulation, but then differences in team sizes and in the experience of team members become very important in comparing what has been achieved by different participants. The third competition, like those before it, did not explore these factors, so it is impossible to make judgements about the relative effectiveness of the solutions to these problems offered by each of the hand-coded planning systems. This remains a very important open issue for future competitions to address.

### B.2.1 SHOP2

SHOP2 (Nau, Au, Ilghami, Kuter, Murdoch, Wu, & Yaman, 2003) is a Hierarchical Task Network (HTN) planner. Like most other HTN planners, SHOP2 allows tasks and subtasks to be partially ordered. Thus plans may interleave subtasks from different tasks during expansion of tasks. However, unlike most other HTN planners, SHOP2 generates the steps of each plan in the same order that those steps will later be executed and it can therefore maintain a representation of the current state at each stage in the planning process. This makes it much easier to incorporate substantial expressive power into the HTNs used by SHOP2. For example, they might include axioms, mixed symbolic and numeric computation, or even calls to external programs.

### B.2.2 TALPLANNER

The most successful of the hand-coded planners in 2000, TALPLANNER (Kvarnström & Magnusson, 2003) uses a temporal action logic as a language for describing planning domains and uses control rules that guide the planner in making intelligent choices while constructing plans in a forward search — an idea originally developed in TLPLAN. The rules can act to prune away search branches that are predicted (by the human encoding the rules) to lead to no solutions. Using this idea, it is possible to arrive at a collection of rules that, by examination of a given state, can guide the planner to choose actions so effectively that virtually no search is required at all.

### B.2.3 TLPLAN

TLPLAN (Bacchus & Kabanza, 2000) also uses a temporal logic language to support the construction of control rules to guide plan search. TLPLAN preceded TALPLANNER in its use of this idea. TLPLAN adopts a slightly different approach to the management of tem-poral structure than TALPLANNER, and is also capable of handling metric quantities. The extensions of TLPLAN that allow it to handle time are described by Bacchus and Ady (2001).

## Appendix C. Statistical Techniques

The analysis conducted in this paper makes use of several standard statistical tests. These are the Wilcoxon matched-pairs rank-sum test, the proportion test, the matched-pairs t-test, Spearman's rank correlation test and the rank correlation test for agreement in multiple judgements. For the benefit of readers who are unfamiliar with these tests, we briefly summarise them here. This appendix was constructed using Gopal K. Kanji's *100 Statistical Tests* (1999).

### C.1 The Proportion Test

This test is also known as the binomial distribution test. The test is used to consider the proportion of a sample for which a particular qualitative observation has been made. For example, the proportion of rolls of a die that have come up 6. The test examines how far from the expected proportion is the observed proportion, given an assumed probability for the observation. In its use in this paper, we adopt a null hypothesis that two planners should win with equal likelihood and test the proportion of observed wins for one planner against this hypothesis. If the deviation of observed proportion from expected proportion is sufficiently high, then the null hypothesis can be rejected.

### C.2 The t-test

The t-test is a parametric test: it is founded on an assumption that the underlying population from which the samples are drawn is nearly normally distributed. It is reasonably robust to failures in this assumption, but should be treated with caution as the true distribution deviates from normal. The test considers means of two samples and tests the null hypothesis that the two samples are drawn from populations with the same mean. Variants are available according to what is known about variance in the underlying populations. The t-test is a more conservative version of the Z-test, which relies on the effect confirmed by the Central Limit Theorem that, for large samples, the sampling ditribution of the mean is normal. The t-test can be applied with smaller samples, compensating for the distortion of the distribution that this creates. In this paper, we use a variant of the t-test in which observations are drawn in matched pairs: each element of a pair is a test result conducted under close to identical circumstances, but for a different test subject (in this case, a different planner).

For $n$ pairs of observations, where $d_i$ is the difference for pair $i$ and $\overline{d}$ is the mean difference, the variance, $s$, of the differences is given by:

$$s^2 = \sum_{i=1}^{n} \frac{(d_i - \overline{d})^2}{n-1}$$

If $\overline{x_1}$ and $\overline{x_2}$ are the means of the samples from each of the two populations, then the statistic is:

$$t = \frac{\overline{x_1} - \overline{x_2}}{s/\sqrt{n}}$$

with $n-1$ degrees of freedom.

## C.3 The Wilcoxon Matched-Pairs Rank-Sum Test

The use of ranks releases statistical tests from the parametric assumptions about underlying distributions by replacing actual observed values with their rank within the ordered set of observed values. The Wilcoxon matched-pairs test is analogous to the matched-pairs t-test, but uses the sum of the ranks of the values associated with each of the two test subjects. The pairs are ordered according to the absolute values of their differences and then the sum of the ranks of the positive values is compared with the sum of the ranks of the negative values. If the two subjects exhibit no particular pattern in their relative behaviours then the positive and negative values should be distributed roughly evenly through the ranks and thus the rank-sums should be approximately equal. A distortion between the rank-sums indicates that one or other subject has a consistently superior performance over the other.

The test is defined as follows. Given a collection of $n$ pairs of data items, the differences between the pairs are found and ranked according to absolute magnitude. The sum of the ranks is then formed for the negative and positive differences separately. $T$ is the smaller of these two rank sums. For sufficiently large samples the following value is approximately normally distributed:

$$\frac{n(n+1)/4 - T}{\sqrt{n(n+1)(2n+1)/24}}$$

## C.4 Spearman's Rank Correlation Test

This is a test for correlation between a sequence of pairs of values. Using ranks eliminates the sensitivity of the correlation test to the function linking the pairs of values. In particular, the standard correlation test is used to find linear relations between test pairs, but the rank correlation test is not restricted in this way.

Given $n$ pairs of observations, $(x_i, y_i)$, the $x_i$ values are assigned a rank value and, separately, the $y_i$ values are assigned a rank. For each pair $(x_i, y_i)$, the corresponding difference, $d_i$ between the $x_i$ and $y_i$ ranks is found. The value $R$ is:

$$R = \sum_{i=1}^{n} d_i^2$$

For large samples the test statistic is then:

$$Z = \frac{6R - n(n^2 - 1)}{n(n+1)\sqrt{n-1}}$$

which is approximately normally distributed.

## C.5 Rank Correlation Test for Agreement in Multiple Judgements

This tests the significance of the correlation between $n$ series of rank numbers, assigned by $n$ judges to $K$ subjects. The $n$ judges give rank numbers to the $K$ subjects and we compute:

$$S = \frac{nK(K^2 - 1)}{12}$$

and $S_D$, the sum of squares of the differences between subjects' mean ranks and the overall mean rank. Let:

$$D_1 = \frac{S_D}{n}, D_2 = S - D_1, S_1^2 = \frac{D_1}{K-1}, S_2^2 = \frac{D_2}{K(n-1)}$$

The test statistic is:

$$F = \frac{S_1^2}{S_2^2}$$

which follows the $F$ distribution with $K-1, K(n-1)$ degrees of freedom.